%% file: iclr2025_conference.tex
\documentclass{article} % For LaTeX2e
\usepackage{iclr2025_conference,times}

\input{math_commands.tex}

\usepackage[utf8]{inputenc} % allow utf-8 input
\usepackage[T1]{fontenc}    % use 8-bit T1 fonts
\usepackage{hyperref}       % hyperlinks
\usepackage{url}            % simple URL typesetting
\usepackage{booktabs}       % professional-quality tables
\usepackage{amsfonts}       % blackboard math symbols
\usepackage{nicefrac}       % compact symbols for 1/2, etc.
\usepackage{microtype}      % microtypography
\usepackage{xcolor}         % colors
\usepackage{natbib}

\usepackage{graphicx}
\usepackage{tikz}
\usepackage{comment}
\usepackage{amsmath,amssymb} % define this before the line numbering.
\usepackage{color}
\usepackage{graphicx}
\usepackage{epsfig}
\usepackage{multirow}
\usepackage{algorithm} 
\usepackage{algorithmic}
\usepackage{pifont}

\title{ComPC: Completing a 3D Point Cloud with 2D Diffusion Priors}

\author{Tianxin Huang, Zhiwen Yan, Yuyang Zhao, Gim Hee Lee \\
School of Computing, National University of Singapore University\\
\texttt{\{huangtx, gimhee.lee\}@nus.edu.sg}
}
\iclrfinalcopy % Uncomment for camera-ready version, but NOT for submission.
\begin{document}

\maketitle

\begin{abstract}
% 3D point cloud completion is designed to recover complete shapes from partially observed point clouds. 
3D point clouds directly collected from objects through sensors are often incomplete due to self-occlusion. Conventional methods for completing these partial point clouds rely on manually organized training sets and are usually limited to object categories seen during training.
In this work, we propose a test-time framework for completing partial point clouds across unseen categories without any requirement for training.
Leveraging point rendering via Gaussian Splatting, we develop techniques of Partial Gaussian Initialization, Zero-shot Fractal Completion, and Point Cloud Extraction that utilize priors from pre-trained 2D diffusion models to infer missing regions and extract uniform completed point clouds.
Experimental results on both synthetic and real-world scanned point clouds demonstrate that our approach outperforms existing methods in completing a variety of objects. Our project page is at \url{https://tianxinhuang.github.io/projects/ComPC/}.
\end{abstract}

\section{Introduction}
3D point clouds have always been an important perceptual approach for the physical world, finding extensive use in various applications such as SLAM~\citep{cadena2016past} or 3D detection~\citep{geiger2013vision,reddy2018carfusion}. However, point clouds are often captured from specific camera viewpoints~\citep{yuan2018pcn,kasten2024point} in real applications, which may lead to the incompleteness of collected points due to the self-occlusion. 
Robust completion for partial point clouds can greatly reduce the cost for data collection, and are useful for subsequent 3D perception. 

\begin{figure*}[t]
	%%\vskip -0.3in
	\begin{center}
		\scalebox{0.9}{
			\centerline{\includegraphics[width=\linewidth]{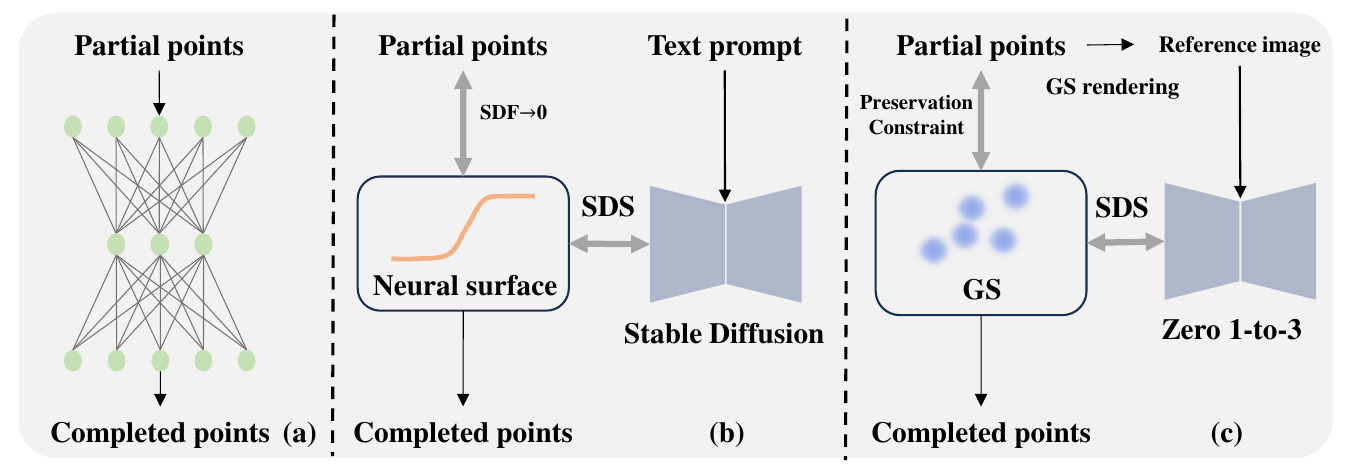}}
		}
		\vskip -0.1in
	\caption{Different point cloud completion methods. (a) Existing network-based completion methods; (b) Test-time SDS-complete~\citep{kasten2024point} with text prompts to guide Neural surface for completion; (c) Our method based on 3D Gaussian Splatting (GS) guided by the diffusion model from Zero 1-to-3~\citep{liu2023zero} conditioned on the reference image rendered from partial points.}
	\label{pic:intro}
	\end{center}
	\vskip -0.3in
\end{figure*}

As illustrated in Fig.~\ref{pic:intro}-(a), most existing completion methods~\citep{yuan2018pcn,zhao2021point,zhou2022seedformer,yu2023adapointr} adopt well-designed deep neural networks to directly generate complete point clouds from partial ones. These methods are usually trained on specific point cloud datasets~\citep{yuan2018pcn,yu2023adapointr} and demonstrate outstanding performances on their respective test sets. 
However, they face challenges in handling data that differs from what they were trained on, such as unseen object categories or real-world scans. This limitation significantly hinders the practical deployment of these point cloud completion methods.
% However, their effectiveness is often compromised by the lack of diversity in the training datasets,

Leveraging the impressive capabilities of 2D diffusion models~\citep{rombach2022high,saharia2022photorealistic,ho2020denoising}, SDS-complete~\citep{kasten2024point} firstly propose a test-time point cloud completion methods utilizing text-to-3D generative models~\citep{poole2022dreamfusion,wang2023score}. 
As shown in Fig.~\ref{pic:intro}-(b), this method optimizes a \textcolor{black}{Neural} surface~\citep{yariv2021volume} guided %from
by Score Distillation Sampling (SDS)~\citep{poole2022dreamfusion}  of the text-conditioned Stable Diffusion~\citep{rombach2022high}. 
The Neural surface, modeled as a Signed Distance Field (SDF) following \textcolor{black}{VolSDF}~\cite{yariv2021volume}, incorporates the geometric details from the partial points by setting their SDF values to zero. The completed points are then generated from the optimized surface for assessment. 
By tapping into the extensive 2D knowledge provided by diffusion models, SDS-complete~\citep{kasten2024point} manages to achieve significantly robust point cloud completion without any training on specific training sets.
% improved robustness in point cloud completion compared to previous network-based methods.
% Capitalized on the abundant 2D priors from 2D diffusion models~\citep{rombach2022high}, \citep{kasten2024point} achieves much better robustness completion performances over aforementioned network-based methods.
However, a notable limitation of the method proposed by SDS-complete~\citep{kasten2024point} is its dependency on manually created text prompts for each point cloud to guide the completion. This requirement can encounter a challenge in real-world applications, where providing detailed and accurate text descriptions for incomplete point clouds is not always feasible. 

% Furthermore, the process of optimizing the NeuS surfaces that is integral to this method is time-intensive.

% Luckily, Zero 1-to-3~\citep{liu2023zero} can 
%Regarding the issues above
In view of the above-mentioned issues, we propose a novel test-time point cloud completion framework that eliminates the need for any extra manually provided information such as text descriptions.
\label{sec:assumption}
% As discussed in PCN~\citep{yuan2018pcn} and SDS-complete~\citep{kasten2024point}, existing completion methods concentrate mainly on point clouds incomplete due to self-occlusion. These point clouds would be almost the same as the completed point clouds when observed from a certain reference viewpoint, such as the viewpoint of the LiDAR or depth scanner. 
% According to Amodal perception~\citep{}, the whole physical structure can be perceived when only parts of it affect the sensory receptors.
\textcolor{black}{As discussed in PCN~\citep{yuan2018pcn} and SDS-complete~\citep{kasten2024point}, existing completion methods concentrate mainly on point clouds incomplete due to self-occlusion, which means that these point clouds often appear nearly complete from at least one viewpoint.
Inspired by the amodal perception~\citep{lehar1999gestalt,breckon2005amodal}, we aim to complete a point cloud by utilizing the observation from a reference viewpoint that provides the most complete view of the point cloud.}
% we explore to complete a point cloud with its observation from a reference viewpoint where the point cloud can be most completely observed.
% In this work, we explore to complete a point cloud with its observation under such a viewpoint.

As illustrated in Fig.~\ref{pic:intro}-(c), we estimate such a viewpoint and acquire a reference image of the partial point cloud.
\textcolor{black}{Inspired by the capability of novel view synthetic diffusion model, e.g., Zero 1-to-3~\citep{liu2023zero}, we propose to use the reference image as a condition for guidance from the diffusion model to infer the missing regions.
Utilizing 3D Gaussian Splatting (GS)~\citep{kerbl20233d}, which can render 2D images from discrete 3D Gaussians initialized from point clouds, we can effectively render the reference image. This approach also allows us to incorporate 2D diffusion priors into the process of modifying 3D geometry. Consequently, we can complete the missing regions by optimizing the 3D Gaussians with guidance from the 2D diffusion model.}
% Since 3D Gaussian Splatting(GS)~\citep{kerbl20233d} can render 2D images from discrete 3D Gaussians initialized from point clouds, it can be naturally used to render the reference image and introduce 2D diffusion priors for 3D geometry modification.
% Therefore, we propose to complete the missing regions by optimizing the 3D Gaussians through guidance from the 2D diffusion model.
% Inspired by the capability of Zero 1-to-3~\citep{liu2023zero} to \textcolor{black}{synthesize multi-angle novel views from a single viewpoint}, we use this reference image as a condition for guidance from the diffusion model of Zero 1-to-3~\citep{liu2023zero} to optimize 3D Gaussians.
% Illustrated in Fig.~\ref{pic:intro}-(c), our initial step involves the colorization of point clouds through 3D Gaussian Splatting (GS), generating a colored and most completed observation of partial points.
% As 
% % with minimal self-occlusion. This observation will be the basis of subsequent completion.
% Inspired by the capability of Zero 1-to-3~\citep{liu2023zero} to infer global geometry from a single viewpoint, we optimize 3D Gaussians with the guidance from Zero 1-to-3 conditioned on the observed results from point clouds colorization. 
% The missing parts are then completed by the optimization of 3D Gaussians guided by 2D diffusion priors.
Moreover, we propose Preservation Constraint to maintain the geometric integrity of partial point clouds. The completed point clouds would be finally acquired from the 3D Gaussian centers.

Our \textbf{main contributions} can be summarized as below:
\begin{itemize}
    % \item We present a zero-shot framework for completing point clouds, leveraging 2D priors from the diffusion model through the technique of 3D Gaussian Splatting;
    \item We propose the Partial Gaussian Initialization to generate a reference image for partial points, which is observed from an estimated reference viewpoint;
    % , where \hl{the partial point cloud can be most completely observed;}
    \item Based on the reference image, we develop the Zero-shot Fractal Completion to complete the missing regions by introducing 2D diffusion priors;
    \item We propose Point Cloud Extraction to extract uniform point clouds from 3D Gaussians;
    \item Through comprehensive evaluation across various data, we demonstrate that our approach surpasses conventional completion methods in handling both synthetic and real-world scanned point clouds.
\end{itemize}

% To achieve perception for 3D world, it is important to 

\section{Related Works}

\subsection{3D Generation via 2D priors}
Since the notable success of 2D diffusion models in text-to-image generation~\citep{rombach2022high,saharia2022photorealistic,ho2020denoising}, text-to-3D and image-to-3D generation have attracted the attention of an increasing number of researchers. 
% However, existing 3D naive diffusion models~\citep{jun2023shap,nichol2022point,gupta20233dgen,lorraine2023att3d,zhang20233dshape2vecset} \textcolor{black}{are }
% are limited by the lack of large-scale 3D datasets, which can usually only be applied to objects from one or a few categories. 
%Single image reconstruction methods 
% In comparison, the abundance and accessibility of 2D data make it feasible to train more universal diffusion models.
To achieve robust and generalizable 3D generation, researchers propose to lift 2D priors for 3D generation~\citep{poole2022dreamfusion,wang2023score,mohammad2022clip,michel2022text2mesh}.
These works usually optimize specific 3D representations by guidance from 2D diffusion models under different viewpoints, where the guidance is calculated with Score Distillation Sampling (SDS)~\citep{poole2022dreamfusion} through rendered images. \textcolor{black}{Score Distillation Sampling (SDS) guides a target model (e.g., NeRF) by using gradients from a pre-trained diffusion model. This aligns the target model's output with the diffusion model's learned distribution, enabling high-quality generation in specialized domains.}

Zero 1-to-3~\citep{liu2023zero} achieve remarkable 3D generation quality by using SDS guidance from their pre-trained novel view synthesis diffusion model explicitly conditioned on the reference image and camera transformation. \textcolor{black}{Conditioned on a single image, Zero 1-to-3 predicts an image consistent with plausible 3D shapes for any given camera pose.}
However, its reliance on NeRF representation leads to prolonged optimization times.
% Dreamgaussian~\citep{tang2023dreamgaussian} optimizes Gaussian Splatting(GS)~\citep{kerbl20233d} representation through SDS guidance from Zero 1-to-3~\citep{liu2023zero}, which achieve both relatively high quality and acceptable optimization time. 
\textbf{3D Gaussian Splatting (GS)}~\citep{kerbl20233d} is an efficient 3D representation that encodes both geometrical and appearance information using a set of 3D Gaussians. Each Gaussian is defined by attributes such as 3D coordinates, scaling, opacity, rotation, and spherical harmonics parameters. By optimizing these attributes, information from 2D images can be incorporated into the Gaussians, enabling efficient novel-view rendering.
Dreamgaussian~\citep{tang2023dreamgaussian} offers a solution by optimizing 3D Gaussians through SDS from Zero 1-to-3, achieving a balance between high-quality outputs and acceptable optimization durations.

Motivated by Dreamgaussian, we recognize the potential of GS to refine 3D coordinates of Gaussian centers using guidance from 2D diffusion models. 
This insight presents an opportunity to apply 2D diffusion priors to tasks related to 3D point clouds, such as point cloud completion.
% through specific 3D representations such as Nerf or Gaussian 
% However, generation based on point clouds remains unexplored in existing works. Fortunately, benefited from the differentiable point rendering of Gaussian Spalatting (GS), point clouds can 

\subsection{Point Cloud Completion}
Point cloud completion aims to recover completed point clouds from partial input point clouds. 
Ever since PCN~\citep{yuan2018pcn} firstly applied deep neural networks to predict complete point clouds from partial inputs, numerous advancements~\citep{zhang2020detail,xie2020grnet,huang2020pf,yu2021pointr,wang2020cascaded,xiang2022snowflake,wen2021pmp} have been made to enhance the accuracy of point cloud completion by altering network architectures. For example, GRNet~\citep{xie2020grnet} converts point clouds into grid formats and employs 3D CNNs for predicting the completed structures, while PFNet~\citep{huang2020pf} adopts a fractal approach to better preserve existing shape details. 
\textcolor{black}{The Fractal approach focuses on predicting only the missing regions of point clouds, preserving existing details by retaining the shapes from the partial input.}
RFNet~\citep{huang2021rfnet} utilizes a differentiable layer to merge existing geometrical details from partial point clouds into completed results.
% new geometrical details into existing partial point clouds. 
\textcolor{black}{More recent approaches~\citep{wang2024pointattn,Zhu_2023_ICCV,li2023proxyformer,yu2021pointr,xiang2022snowflake,zhou2022seedformer,yu2023adapointr,yan2022shapeformer}} integrate carefully-designed transformers to improve completion accuracy by considering broader geometric relationships.
\textcolor{black}{DiffComplete~\cite{chu2024diffcomplete} is a diffusion-based model for 3D shape completion, leveraging probabilistic modeling to predict missing parts of 3D shapes while preserving structural coherence and diversity.}

However, the effectiveness of these point cloud completion methods diminishes when applied to data that differ from their training sets, such as point clouds from unseen categories or other datasets.
SDS-complete~\citep{kasten2024point} proposed a test-time completion framework that employs \textcolor{black}{VolSDF}~\citep{yariv2021volume} for rendering, drawing on priors from pre-trained text-to-image 2D diffusion models~\citep{rombach2022high}. This approach maintains the original shapes by constraining the Signed Distance Field (SDF) values of the partial inputs. 
Yet, this strategy's reliance on text-to-image diffusion models for guidance necessitates well-defined text prompts for each partial point cloud, which may not be practical in real-world applications. 
Moreover, the optimization of SDS-Complete is quite time-consuming, which may take more than 1000 minutes for one point cloud.
% Moreover, the rendering based on NeuS is relatively time-consuming.

In this study, we propose to leverage 3D Gaussian Splatting (GS)~\citep{kerbl20233d} to bridge point clouds with priors from 2D diffusion models. 
By generating a reference image of the partial point cloud to serve as a condition for guidance from Zero 1-to-3~\citep{liu2023zero}, our method can extract uniform and completed point clouds from the 3D Gaussian centers. Since our method exclusively utilizes information gathered from the incomplete point cloud for completion, it eliminates the need for any additional manually specified prompts for each point cloud. 
Due to the efficient rendering from 3D GS, and stronger priors from Zero 1-to-3, our method can achieve much higher optimization efficiency than SDS-Complete~\citep{kasten2024point}.

\section{Methodology}
\label{sec:metho}
\begin{figure}[t]
	% \vskip -0.3in
	\begin{center}
		\scalebox{0.9}{
			\centerline{\includegraphics[width=\linewidth]{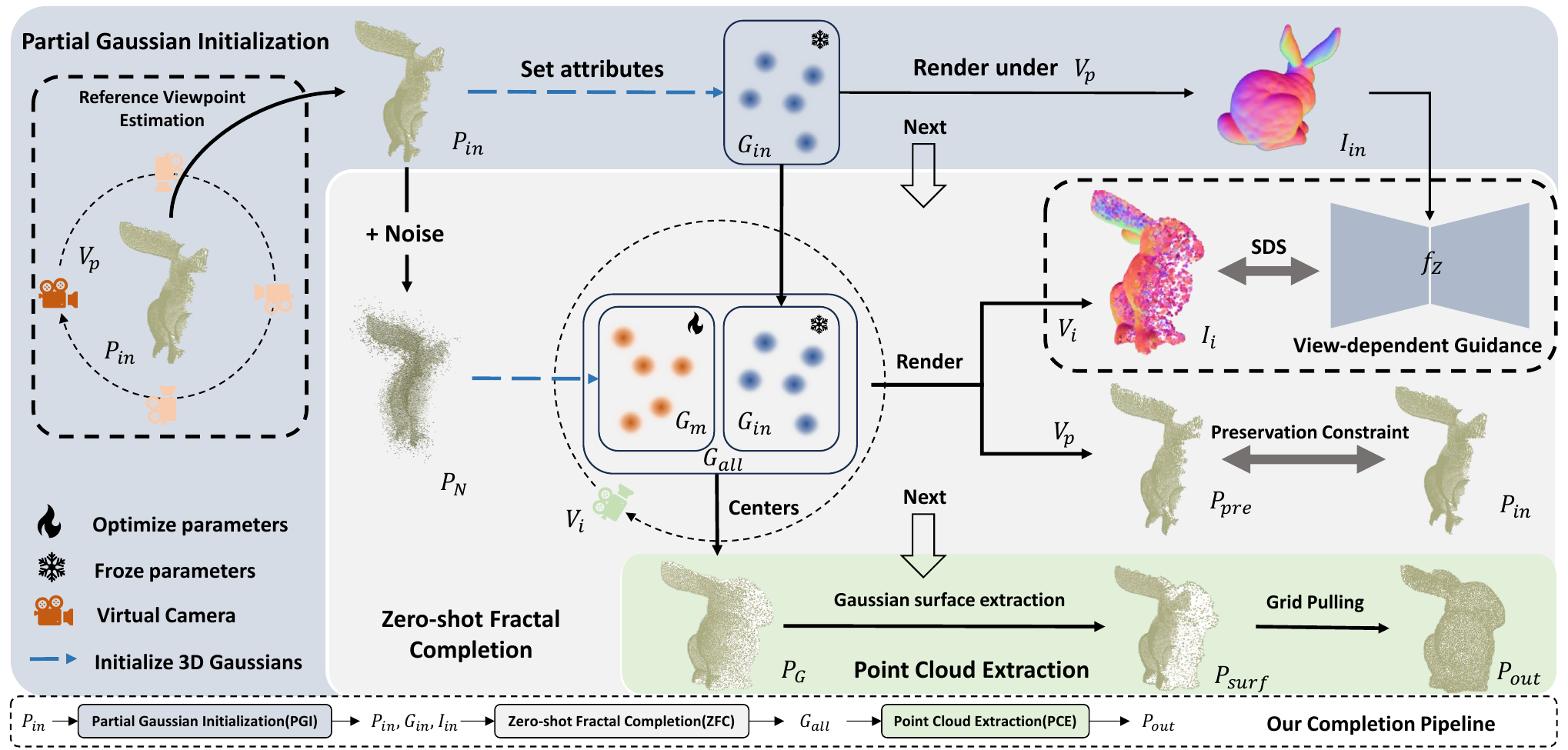}}
		}
		\vskip -0.05in
		% \caption{Illustration of our framework. In Partial Gaussian Initialization, Reference Viewpoint Estimation estimates a camera pose $V_p$ where $P_{in}$ can be most completely observed. We initialize 3D Gaussians $G_{in}$ from $P_{in}$ and render the reference image $I_{in}$ under $V_p$. In Zero-shot Fractal Completion, 3D gaussians $G_m$ begins with an initialization using noisy $P_{N}$ and undergoes optimization guided by view-dependent guidance from the pre-trained diffusion model $f_Z$ in Zero 1-to-3~\citep{liu2023zero} based on a randomly chosen camera pose $V_i$. Additionally, it incorporates a Preservation Constraint computed with respect to $V_p$. $G_{in}$ is mixed with $G_m$ to form $G_{all}$, introducing the partial geometry from $P_{in}$. After optimization, we use Point Cloud Extraction to extract surface points $P_{surf}$ from centers of $G_{all}$, and convert $P_{surf}$ into uniform $P_{out}$ with Grid Pulling.
  % % $G_{in}$ includes 3D Gaussians initialized from $P_{in}$. We render the reference image $I_{in}$ under $V_p$ following DreamGaussian~\citep{tang2023dreamgaussian}.
  % }
  \caption{\textcolor{black}{Illustration of our framework. \ding{172}In Partial Gaussian Initialization (PGI), Reference Viewpoint Estimation estimates a camera pose $V_p$ where $P_{in}$ can be most completely observed. We initialize 3D Gaussians $G_{in}$ from $P_{in}$ and render the reference image $I_{in}$ under $V_p$. \ding{173}In Zero-shot Fractal Completion (ZFC), 3D Gaussians $G_m$ begins with an initialization using noisy $P_{N}$ and undergoes optimization guided by view-dependent guidance from the diffusion model $f_Z$ in Zero 1-to-3~\citep{liu2023zero} based on a randomly chosen camera pose $V_i$. Additionally, it incorporates a Preservation Constraint computed with respect to $V_p$. $G_{in}$ is mixed with $G_m$ to form $G_{all}$, introducing the partial geometry from $P_{in}$. \ding{174}After ZFC, we use Point Cloud Extraction (PCE) to extract surface points $P_{surf}$ from centers of $G_{all}$, and convert $P_{surf}$ into uniform $P_{out}$ with Grid Pulling.}
  % $G_{in}$ includes 3D Gaussians initialized from $P_{in}$. We render the reference image $I_{in}$ under $V_p$ following DreamGaussian~\citep{tang2023dreamgaussian}.
  }
		\label{pic:framework}
	\end{center}
	\vskip -0.25in
\end{figure}

As shown in Fig.~\ref{pic:framework}, the whole completion process is composed of Partial Gaussian Initialization (PGI), Zero-shot Fractal Completion (ZFC), and Point Cloud Extraction (PCE). For the given partial point cloud $P_{in}$, we firstly transform it into colorized reference image $I_{in}$ and 3D Gaussians $G_{in}$ with Partial Gaussian Initialization. 
Subsequently, $I_{in}$ and $G_{in}$ are introduced to Zero-shot Fractal Completion to acquire 3D Gaussians $G_{all}$ with the completed shape. Specifically, we use $I_{in}$ to guide the optimization of 3D Gaussians $G_m$ by borrowing priors from the 2D diffusion model in Zero 1-to-3~\citep{liu2023zero}.
Finally, we extract uniform completed point clouds $P_{out}$ from the centers of $G_{all}$ with Point Cloud Extraction. 
\textbf{Please note that the completion is mainly achieved by optimizing 3D Gaussian parameters in $G_{m}$}, without networks as \cite{yuan2018pcn}.
% More details are presented below.

% completed point cloud $P_{out}$. 
% We use $I_{in}$ to guide the optimization of 3D Gaussians $G_m$ for completion, and finally acquire the completed point clouds $P_{out}$ from the centers of Gaussians $G_{in}$ and $G_m$. Details are presented below.

% \subsection{Preliminary: 3D Gaussian Splatting}
% In this section, we provide a brief overview about 3D Gaussian Splatting~\citep{kerbl20233d}.
% 3D Gaussian Splatting~\citep{kerbl20233d} is an efficient 3D representation, which describes 3D geometrical and appearance information with the combination of multiple 3D Gaussians. Each 3D Gaussian consists of attributes including: 3D coordinates, scaling, opacity, rotation, and spherical harmonics parameters. During optimization, attributes of 3D Gaussians are optimized to fit the provided images. During inference, we can render synthetic images under novel views from 3D Gaussians efficiently. More details can be found in \citep{kerbl20233d}.

\subsection{Partial Gaussian Initialization}
\textcolor{black}{Following the definition of point cloud completion task by PCN~\citep{yuan2018pcn}, only \textbf{3D coordinates} are provided as input to infer the complete geometry.}
To introduce priors from pre-trained 2D diffusion models, we use 3D Gaussian Splatting (GS) to achieve differentiable rendering from 3D point clouds to 2D images.
In Partial Gaussian Initialization, we firstly estimate a reference camera pose $V_p$ with the Reference Viewpoint Estimation. 
Then, we initialize 3D Gaussians $G_{in}$ from the incomplete point cloud $P_{in}$. A reference image $I_{in}$ for subsequent completion would be rendered from $G_{in}$ under the pose $V_p$.
$G_{in}$ is frozen to preserve geometrical characteristics of $P_{in}$.

\noindent\textbf{Reference Viewpoint Estimation.}
For any point cloud to be completed, we first determine an reference camera pose $V_p$, that captures its most completed observation. The completion process then builds upon this observation. Since the incomplete point cloud $P_{in}$ typically spans across a surface, its most complete view is characterized by minimal self-occlusion and closeness to the camera.

Considering the potential occlusion of rear Gaussians by those in the foreground during rendering,
we implement a filter $h(G_{in}, V_n)$ to identify the indices of the frontmost 3D Gaussians in $G_{in}$ from the camera pose $V_n$.
\label{surfilter}
Given that the centers of $G_{in}$ are anchored to $P_{in}$, we can estimate $V_p$ by minimizing: %the following equation:
\begin{equation}
    V_p = \arg\min_{V_n} \operatorname{CD}(P_{in}[h(G_{in}, V_n)], P_{in}) + w_0 \cdot \operatorname{Depth}(P_{in}, V_n),
\end{equation}
where $\operatorname{CD}(\cdot,\cdot)$ is the Chamfer Distance~\citep{fan2017point} to measure shape differences between two point clouds. $\operatorname{Depth}(P_{in}, V_n)$ calculates the mean depths of $P_{in}$ observed from the camera at pose $V_n$ for regularization, and $w_0$ is a weighting factor to ensure balance. 
% \textcolor{black}{Please note that this estimation is designed to acquire an observation as complete as possible, instead of estimating the absolutely correct camera pose.}
For this study, we estimate $V_p$ by examining 5,000 camera positions uniformly distributed around the partial point cloud.

% \vspace{2mm}
\noindent\textbf{Gaussian Attributes Setting.}
\label{newGS}
% Upon estimating the reference camera pose $V_p$, we render a depth map $D_{in}$ and a reference image $I_{in}$ from the 3D Gaussians $G_{in}$. 
Upon estimating the reference camera pose $V_p$, we render a reference image $I_{in}$ from 3D Gaussians $G_{in}$ initialized from partial point cloud $P_{in}$. 
To render a characteristic reference image, we make a few modifications to the original 3D Gaussians: 
% 1) Considering point clouds are observed as multiple equal size spheres, we set the scaling of all 3D Gaussians to a single shared scalar value to keep the shape of Gaussians consistent as points. The scaling attributes $G_{in}^s = \frac{1}{|P_{in}|} \sum Neighbor(P_{in})$, where $Neighbor$ denotes the nearest neighbor distance of each point in $P_{in}$.

1) The opacity $G_{in}^o$ for all 3D Gaussians within $G_{in}$ is set to a constant value of 1. This step ensures that Gaussians representing all partial points are nearly opaque and clearly visible during rendering.

2) The color $G_{in}^c$ are set as scaled normal map as: $G_{in}^c = (1+\mathcal{N}(P_{in}))/2$, where the normal vectors $\mathcal{N}(P_{in})$ are estimated with Open3d~\citep{zhou2018open3d}. 
We scale them from $-1 \sim 1$ to $0 \sim 1$.

\subsection{Zero-shot Fractal Completion}
\label{sec:zfc}
Zero-shot Fractal Completion (ZFC) aims to introduce priors to transform $G_{in}$ with the partial shape into $G_{all}$ with the completed shape.
% complete partial shapes from $G_{in}$ into completed $G_{all}$. 
% reconstruct complete and uniform point clouds $P_{out}$ from the colorized reference image $I_{in}$ and 3D Gaussians $G_{in}$. 
As illustrated in Fig.~\ref{pic:framework}, ZFC optimizes 3D Gaussians $G_m$ for completion and is guided by the View-dependent Guidance and the Preservation Constraint. 
% The initially completed point cloud $P_{surf}$, extracted from Gaussian centers $P_G$ through Gaussian surface extraction, is transformed into a uniform output point cloud $P_{out}$ by the Grid Pulling module.

% \vspace{2mm}
\noindent\textbf{Modification for 3D Gaussians.}
% 1) We also adjust the scaling of 3D Gaussians as discussed in Sec.~\ref{newGS} to ensure the shapes of 3D Gaussians are consistent as points in point clouds.
1) Considering point clouds are observed as multiple equal size spheres, we set the scaling of all 3D Gaussians to a single shared scalar value to keep the shape of Gaussians consistent as points. 
To better cover the space around the partial point cloud $P_{in}$, we create noised $P_{N} = P_{in}+ \mathcal{N}(0, \sigma_n^2)$ for the initialization of $G_m$.
The scaling attribute of $G_m$ is initialized as $G_{m}^s = \frac{1}{|P_{N}|} \sum Neighbor(P_{N})$ from the noisy $P_N$ as shown in Fig.~\ref{pic:framework}, where $Neighbor$ denotes the nearest neighbor distance of each point in $P_{N}$.

\begin{figure*}[t]
	%%\vskip -0.3in
	\begin{center}
		\scalebox{0.6}{
			\centerline{\includegraphics[width=\linewidth]{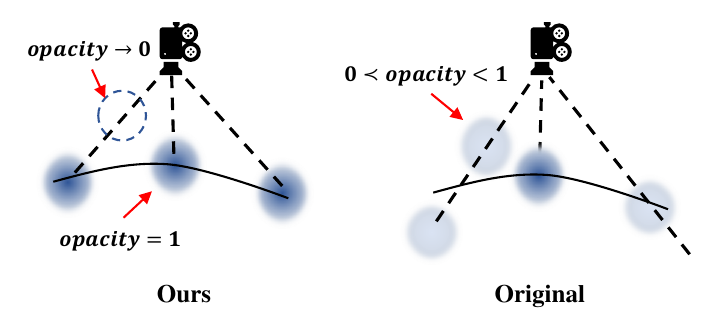}}
		}
		\vskip -0.1in
		\caption{Differences between our binarized opacity and original continuous opacity. $\prec$ denotes smaller but not approaching. }
		\label{pic:opacity}
	\end{center}
	\vskip -0.2in
\end{figure*}

2) Furthermore, as demonstrated in Fig.~\ref{pic:opacity}, the original approach to opacity can lead to a dispersion of Gaussian centers around the actual surface due to the range of opacities $0 \prec opacity < 1$ used in rendering. To address this problem, we apply a differentiable quantization~\citep{huang20223qnet} for Gaussian opacity to binarize the values. For 3D Gaussians $G_m$ with original opacity $G^o_m$, the binarization is implemented as follows:
\begin{equation}
\label{eq:binary}
    G^o_m = f_{stop}(\operatorname{round}(G^o_m)-G^o_m)+G^o_m,
\end{equation}
where
\[
\operatorname{round}(G^o_m) = 
\begin{cases} 
1 & \text{if } G^o_m > 0.5, \\
\delta & \text{otherwise}.
\end{cases}
\]
with $f_{stop}(\cdot)$ designed to halt gradient propagation. Here, the forward propagation result of Eq.~\ref{eq:binary} is $\operatorname{round}(G^o_m)$, while the gradient during backpropagation is calculated based on $G^o_m$. $\delta$ is a predefined small constant set to 0.01 in this work because lower opacity may make the Gaussians hard to optimize.
Consequently, 3D Gaussians with $G^o_m \rightarrow 1$ cluster near the surface as shown in Fig.~\ref{pic:opacity}, while those with $G^o_m \rightarrow 0$ will be considered noise and excluded in subsequent processing.

% Zero-shot Fractal Completion (ZFC) aims to recover completed and uniform point clouds $P_{out}$ from the colorized $I_{in}$ and $G_{in}$. 
% As shown in Fig.~\ref{pic:framework}, ZFC optimizes 3D Gaussians $G_m$ for completion under the controlling of View-dependent Guidance and Preservation Constraint. Initialized completed point cloud $P_{surf}$ is extracted from Gaussian centers $P_G$ with Gaussian surface extraction and resampled into uniform output point cloud $P_{out}$ by Grid Pulling module.

% In ZFC, we also make modifications on the scaling attributes of 3D Gaussians following Sec.~\ref{newGS}, in order to keep the 3D Gaussian surface consistent with Gaussian centers. 
% %3D Gaussians with $opacity \rightarrow 0$ hardly affects the rendering, $opacity \rightarrow 1$ Gaussians 
% Furthermore, as shown in Fig.~\ref{pic:opacity}, the original continuous opacity may produce multiple 3D Gaussians with $0 \prec opacity < 1$ for rendering. Gaussian centers in this setting may distribute around instead of exactly on the true surface.
% We introduce a differentiable quantization operation for the opacity attributes to reduce the noises. Given the original opacity of 3D Gaussians $G_m$ as $G^o_m$, we binarized it following \citep{huang20223qnet} as
% \begin{equation}
% \label{eq:binary}
%     G^o_m = f_{stop}(round(G^o_m)-G^o_m)+G^o_m,
% \end{equation}
% \[
% round(G^o_m) = 
% \begin{cases} 
% 1 & \text{if } G^o_m > 0.5, \\
% \delta & \text{otherwise}
% \end{cases}
% \]
% where $f_{stop}$ means to stop the gradient propagation. In this case, the forward propagation result of Eq.~\ref{eq:binary} would be $round(G^o_m)$, while gradient is calculated according to $G^o_m$ during back propagation. $\delta$ is a pre-defined small constant.
% In this work, we set $\delta$ as 0.01 because lower opacity may make the Gaussians hard to optimize.
% 3D Gaussians with $G^o_m \rightarrow 1$ will gather near the surface as shown in Fig.~\ref{pic:opacity}. 
% Noisy Gaussians with $G^o_m \rightarrow 0$ have hardly contributions and will be removed in the Gaussian Surface Extraction.

% \vspace{2mm}
\noindent \textbf{View-dependent Guidance.}
\label{sec:guide}
To complete the missing regions, we leverage 2D diffusion priors from Zero 1-to-3~\citep{liu2023zero} due to its capability to deduce the unseen regions based on available imagery.
As illustrated in Fig.~\ref{pic:framework}, we utilize the reference image $I_{in}$ from Partial Gaussian Initialization to derive the SDS guidance~\citep{poole2022dreamfusion} based on image $I_i$ rendered with Gaussian Splatting in a randomly selected viewpoint \(V_i\), referred to as View-dependent guidance. Defining \(\epsilon_{f_Z}\) as the noise anticipated by the 2D diffusion model \(f_Z\) with \(t\) and \(\epsilon\) indicating the time step and standard noise, respectively, the SDS guidance is calculated as:
\begin{equation}
\nabla_{G_{all}} L_{SDS} = \mathbb{E}
_{t, \epsilon}[(\epsilon_{f_Z}(I_{i};I_{in}, V_i, t)-\epsilon)\frac{\partial I_{i}}{\partial G_{all}}].
\end{equation}
For the task of point cloud completion, we adopt a fractal approach as discussed in PFNet~\citep{huang2020pf}, focusing on optimizing only \(G_m\) within \(G_{all}\) for reconstructing missing regions, while \(G_{in}\) remains unchanged to conserve the original geometric characteristic of the partial point clouds \(P_{in}\).
Additionally, to manage the scaling \(G^s_{m}\) of 3D Gaussians $G_m$ during optimization, we implement a regularization with a weighting factor of $w_1$:
\begin{equation}
    L_{mreg} = w_1 \cdot |G^s_{m}|.
\end{equation}

\noindent \textbf{Preservation Constraint.}
To maintain the geometric shapes of the initial partial point clouds, we introduce Preservation Constraint aimed at reducing the shape differences between the partial point cloud $P_{in}$ and Gaussian center coordinates $P_{pre}$ acquired from the partial observation of 3D Gaussians $G_{all}$ under $V_p$.
% $P_{pre}$ from the aforementioned estimated camera pose $V_p$, where $P_{in}$ appears most complete. 
Utilizing the surface filter $h(\cdot,\cdot)$ presented in Sec.~\ref{surfilter}, and considering $G_{all}$ as the combined set of $G_m$ and $G_{in}$ with $P_G[\cdot]$ representing the centers of $G_{all}$, the observed Gaussians centers would be $P_{pre} = P_G[h(G_{all}, V_p)]$.
The Preservation Constraint is formulated as:
\begin{equation}
    L_{p} = w_2 \cdot \operatorname{CD}(P_{pre}, P_{in}),
\end{equation}
where $\operatorname{CD}(\cdot,\cdot)$ is the Chamfer Distance~\citep{fan2017point}. $w_2$ is the weighting factor. This constraint ensures the alignment of $G_{all}$ with $P_{in}$ when observed from the reference camera pose $V_p$.
% To preserve the geometrical characteristic of input partial point clouds, we propose Preservation Constraint to minimize the shape differences between the partial point cloud $P_{in}$ and projected point cloud $P_{proj}$ under estimated camera pose $V_p$ where $P_{in}$ has the most completed observation. 
% Following the definition of surface filter in Sec.~\ref{surfilter} as $h(\cdot)$. 
% Let $G_{all} = [G_m, G_{in}]$ be the concatenation of $G_m$ and $G_{in}$, $P_G$ be the centers of $G_{all}$, the Preservation Constraint can be defined as
% \begin{equation}
%     L_{p} = w_3 \cdot CD(P_G[h(G_{all}, V_p)], P_{in}),
% \end{equation}
% where $CD$ denotes the Chamfer Distance. Preservation Constraint ensures that the completed point cloud is consistent with partial $P_{in}$ when observed from $V_p$.
% under the most completed observation

\subsection{Point Cloud Extraction}
After the optimization of ZFC, we extract point cloud $P_{out}$ from centers of 3D Gaussians $G_{all}$ with Point Cloud Extraction.
Specifically, we firstly select surface points $P_{surf}$ from Gaussian centers $P_G$ with Gaussian surface extraction. Then, we resample uniform $P_{out}$ from $P_{surf}$ by Grid Pulling.
% \vspace{2mm}
\begin{algorithm}[t]
	\caption{Gaussian Surface Extraction}
	\label{alg:surface}
	\begin{algorithmic}[1]
        \STATE {Input: 3D Gaussians $G_{all}$ and corresponding centers $P_G$, $h(\cdot)$ following Sec.~\ref{surfilter}} 
	\STATE {\bfseries Filtering with opacity:} 
	\STATE Let opacity  of $G_{all}$ be $G^o_{all}$,
        \STATE Effective 3D Gaussian indexes $id_o = G^o_{all}>0.5$,
  
         \STATE {\bfseries Extracting the surface points:} 
         \STATE Set a index list $idx=[~]$, generate $N$ uniform camera poses $V$, 
         \FOR{$i=1$ {\bfseries to} $N$}
         \STATE Adding the first observed Gaussian indexes: $idx.\operatorname{append}(h(G_{all}[id_o], V[i]))$
         \ENDFOR
         \STATE Remove the repeated indexes: $idx = \operatorname{Unique}(idx)$
         \STATE Acquire the surface points: $P_{surf} = P_G[id_o][idx]$

   \vskip -0.2in
	\end{algorithmic}
\end{algorithm}
\begin{figure*}[t]
	\vskip -0.1in
	\begin{center}
		\scalebox{1.0}{
			\centerline{\includegraphics[width=\linewidth]{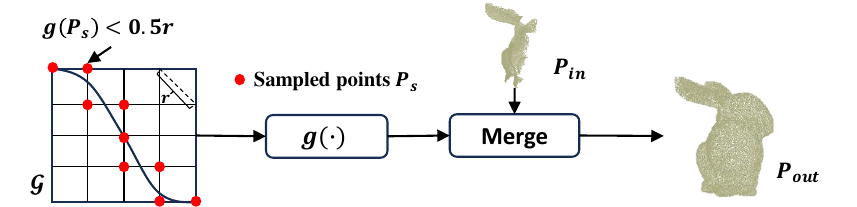}}
		}
		% \vskip -0.1in
		\caption{Illustration of Grid Pulling module. $g(\cdot)$ is a MLP-based SDF learned from the completed point cloud $P_{surf}$. Merge denotes merge layer from \citep{huang2021rfnet}. Given the 3D grids $\mathcal{G}$, $r$ is the diagonal length of a unit grid. Sampled points would be $P_s = \{p \mid g(p)<0.5r, p \in \mathcal{G}\}$.}
		\label{pic:gp}
	\end{center}
	\vskip -0.2in
\end{figure*}

\noindent \textbf{Gaussian Surface Extraction.}
% The optimization of 3D Gaussian centers results in their dispersion around the object's surface, encompassing both interior and exterior regions. 
% The optimization results of 3D Gaussian centers are actually scattered both on the surface and beneath it.
% The optimization results of 3D Gaussian centers are not only on the surface of the shape, but also inside the shape.
The centers of the 3D Gaussians can lie both on and below the surface of the shape after optimization.
%
% Directly using these centers as the completed point cloud is not viable due to this problem.
As a result, it is unsatisfactory to directly use these centers as the complete point cloud. 
To address this issue, we introduce a Gaussian Surface Extraction process to select surface points $P_{surf}$ from the centers of 3D Gaussian $G_{all}$. 
This procedure is detailed in Alg.~\ref{alg:surface}. By adjusting the opacity of all 3D Gaussians to either $\delta$ or 1, we note that Gaussians with minimal opacity $\delta$ hardly contributes to the rendering process. Consequently, our initial step involves filtering $G_{all}$ based on opacity as outlined in Alg.~\ref{alg:surface}. To this end, $G_{all}$ is examined from $N$ uniformly distributed camera positions and $h(\cdot,\cdot)$ is employed to extract the centers of the frontmost visible Gaussians as the surface points $P_{surf}$. We set $N=500$ in this work.
% The optimization results of 3D Gaussian centers are actually distributed around the surfaces, both inside and outside. These 3D Gaussian centers are not appropriate to be directly used as the completed point clouds. Therefore, we propose a Gaussian Surface Extraction operation to extract surface points $P_{surf}$ from 3D Gaussian centers $P_G$. 
% The operation details are presented in Alg.~\ref{alg:surface}.

% As we push the opacity of all 3D Gaussians to 0.01 or 1, Gaussians with low opacity hardly contributes to the rendering. Therefore, we firstly filter the 3D Gaussians $G_{all}$ with opacity in Alg.~\ref{alg:surface}. Then, we observe $G_{all}$ from multiple uniform camera poses, and use $h(\cdot)$ to extract the centers of first observed Gaussians as surface points $P_{surf}$.
% In this work, we propose a Gaussian surface extraction algorithm from 3D Gaussian centers. 
% By simply picking out the center of first Gaussian in each pixel during Gaussian splatting rendering.
\begin{figure*}[t]
	%%\vskip -0.3in
	\begin{center}
		\scalebox{1.0}{
			\centerline{\includegraphics[width=\linewidth]{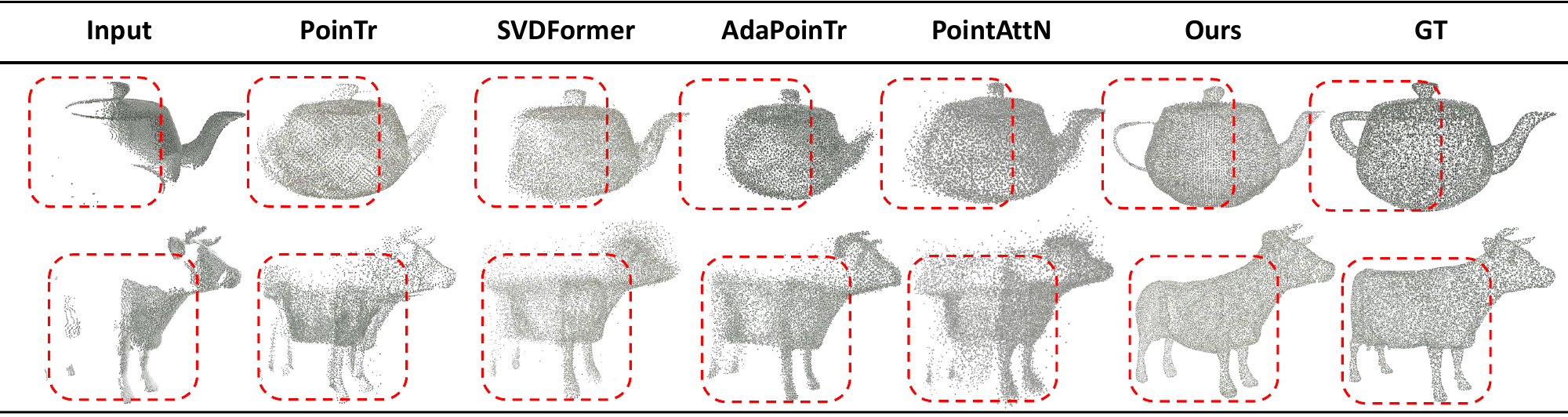}}
		}
		\vskip -0.1in
		\caption{Qualitative comparison on synthetic data.}
		\label{pic:quali_obj}
	\end{center}
	\vskip -0.2in
\end{figure*}
\begin{table}[t]
		\normalsize
		\setlength\tabcolsep{1.2pt}
  \caption{Quantitative comparison on synthetic data. \textbf{Bold} marks the best results.}
		\begin{center}
  % \vskip -0.05in
			\scalebox{0.75}{
				\begin{tabular}{cccccccccccc}
\toprule
Object      & Horse     & MaxPlanck & Armadillo & Cow       & Homer     & Teapot    & Bunny     & Nefertiti & Bimba     & Ogre      & Aver      \\ \cmidrule{2-12}
Metrics    & CD/EMD    & CD/EMD    & CD/EMD    & CD/EMD    & CD/EMD    & CD/EMD    & CD/EMD    & CD/EMD    & CD/EMD    & CD/EMD    & CD/EMD    \\
PoinTr     & 2.75/4.47 & 6.34/6.84 & 3.51/6.07 & 3.13/4.25 & 1.90/4.19 & 3.81/5.12 & 6.39/8.03 & 4.29/5.50 & 5.53/6.73 & 3.41/5.06 & 4.10/5.63 \\
% Snowflake  & 3.34/5.26 & 6.33/7.18 & 3.03/6.13 & \textbf{2.61}/4.27 & 1.89/3.53 & 3.22/4.69 & 6.99/8.81 & 4.15/5.94 & 5.36/7.11 & 3.18/5.46 & 4.01/5.84 \\
SeedFormer & 3.24/5.30 & 6.91/7.62 & 3.28/6.21 & 3.11/4.00 & 2.04/3.52 & 3.41/4.94 & 6.92/9.10 & 4.25/5.78 & 5.63/7.09 & 3.31/5.73 & 4.21/5.93 \\
PointAttN  & 5.25/6.76 & 8.10/8.54 & 5.09/6.65 & 3.73/4.56 & 2.39/3.54 & 5.25/6.36 & 9.35/9.52 & 5.16/5.87 & 8.09/7.52 & 4.80/6.14 & 5.72/6.54 \\
% AnchorFormer  & 3.64/4.63 & 6.07/6.56 & 4.38/5.80 & 2.60/3.57 & \textbf{1.06}/1.99 & 3.60/4.33 & 9.01/8.87 & 4.10/5.20 & 6.73/6.91 & 4.05/5.27 & 4.52/5.31 \\
\textcolor{black}{ShapeFormer}  & 4.17/5.38 & 3.48/4.49 & 3.76/4.68 & 4.53/5.29 & 2.27/2.84 & 2.55/2.86 & 4.52/4.44 & 3.09/3.87 & 5.00/5.85 & 3.39/4.69 & 3.68/4.44 \\
SVDFormer  & 2.70/3.89 & 8.37/6.45 & 4.12/6.53 & 3.55/4.39 & 2.42/3.35 & 5.87/6.08 & 6.59/6.90 & 4.27/5.02 & 5.47/4.91 & 4.59/5.36 & 4.79/5.29 \\
AdaPoinTr  & 4.88/5.45 & 8.60/8.51 & 5.14/5.95 & 3.48/4.53 & 2.28/3.34 & 3.92/4.56 & 9.33/8.87 & 5.54/6.14 & 8.16/7.64 & 4.53/5.41 & 5.59/6.04 \\
Ours       & \textbf{0.96/1.32} & \textbf{1.23/1.53} & \textbf{2.49/4.05} & \textbf{1.45/1.64} & \textbf{1.34}/\textbf{1.76} & \textbf{0.99/1.22} & \textbf{1.43/1.78} & \textbf{1.81/2.20} & \textbf{1.39/1.64} & \textbf{1.22/1.67} & \textbf{1.43/1.88} \\ 
% Ours       & \textbf{1.01/1.36} & \textbf{1.07/1.37} & \textbf{2.00/2.87} & \textbf{1.68/1.74} & \textbf{2.19/2.76} & \textbf{0.96/1.21} & \textbf{1.34/1.65} & \textbf{1.72/2.02} & \textbf{1.42/1.64} & \textbf{1.82/2.44} & \textbf{1.52/1.91} \\ 
\bottomrule
\end{tabular}
			}
		\end{center}
		% \vskip -0.1in
		\label{quan_obj}
		\vskip -0.2in
	\end{table}

\noindent \textbf{Grid Pulling.}
\label{sec:gp}
%In Fig.~\ref{pic:framework}, it's evident 
It is evident in Fig.~\ref{pic:framework} that the density of points in the completed regions of $P_{surf}$ can significantly differ from that in the original partial point clouds. Ideally, we aim for a consistently dense and uniform distribution of points across the entire shape. 
Direct attempts to enhance point density within the Zero-shot Fractal Completion (ZFC) would lead to a substantial increase in computational cost. Inspired by NeuralPull~\citep{ma2020neural}, we introduce a Grid Pulling (GP) module designed to resample points uniformly from initially non-uniform point clouds.

NeuralPull~\citep{ma2020neural} employs a Signed Distance Field (SDF) $g(\cdot)$ to pull randomly sampled points $P_{sam}$
that are often generated by adding noise to $P_{gt}$ as $P_{sam} = P_{gt}+N(0, \sigma_0^2)$ towards the surface defined by the original point cloud $P_{gt}$. 
$\sigma_0$ is the standard deviation for normal distribution $N(0, \sigma_0^2)$.
The pulling operation is defined as: $P_{pull} = P_{sam}-g(P_{sam}) \cdot \nabla g(P_{sam})/\|g(P_{sam})\|_2$. The optimization of $g(\cdot)$ is guided by the Chamfer Distance (CD) as a measure of the distance between $P_{gt}$ and the adjusted points:
\begin{equation}
L_{pull}(P_{sam}, P_{gt}) = \operatorname{CD}(P_{pull}, P_{gt}).
\end{equation}

Leveraging \(P_{surf}\) obtained from Gaussian Surface Extraction, GP module learns an SDF $g(\cdot)$ to align uniformly sampled points around $P_{surf}$ with its surface. Unlike NeuralPull, which optimizes using only noised point clouds, our approach trains $g(\cdot)$ with both noised point clouds $P_{near}=P_{surf}+N(0, \sigma_0^2)$, and $P_{far}$ being randomly sampled within the 3D bounding box encompassing $P_{surf}$. The loss functions are defined as $L_{far} = L_{pull}(P_{far}, P_{surf})$ and $L_{near} = L_{pull}(P_{near}, P_{surf})$.

Additionally, we utilize a merge layer as suggested by \citet{huang2021rfnet} to incorporate geometric details from $P_{in}$ into $P_{pull}$.
% , focusing on the distances between $P_{pull}$ points and their nearest neighbors in $P_{in}$. 
Given the distances from $P_{pull}$ points to their nearest neighbors in $P_{in}$ as $dist = \min_{x \in P_{pull}, \forall y \in P_{in}}\|x-y\|_2$, and corresponding neighbor indexes $idx = \arg\min_{x \in P_{pull}, \forall y \in P_{in}}\|x-y\|_2$, the merge layer $g_m$ outputs a set of merged points: 
\begin{equation}
    g_m(P_{pull}, P_{in}) = e^{-\frac{dist}{\sigma}}P_{in}[idx]+(1-e^{-\frac{dist}{\sigma}})P_{pull},
\end{equation}
where $\sigma$ is a small optimizable variable to decide how much to merge. The corresponding loss would be $L_{mer} = L_{pull}(g_m(P_{pull}, P_{in}), P_{surf}) + w_3 \cdot \|\sigma\|_2$, where $w_3$ is the weighting factor for the regularization of $\sigma$.
The overall training loss for $g(\cdot)$ is then:
\begin{equation}
    L_g = L_{far} + L_{near} + L_{mer}.
\end{equation}

As depicted in Fig.~\ref{pic:gp}, we initialize a $128^3$ 3D grid $\mathcal{G}$ according to the bounding box of $P_{surf}$. Uniform points $P_s$ would be selected by $P_s = \{p \mid g(p)<0.5r, p \in \mathcal{G}\}$. $P_s$ is then pulled to the surface of $P_{surf}$ and combined with $P_{in}$ through merge layer.
The output point clouds would be $P_{out} = g_m(P_s-g(P_s) \cdot \nabla g(P_s)/\|g(P_s)\|_2, P_{in})$. 
As $P_{out}$ is quite dense, we sample it to the specified resolution during comparisons.
\section{Experiments}
\label{sec:datasets}
% To assess the effectiveness of our method, we conduct comparisons on objects derived from both synthetic data and real-world scans. 
% As a test-time point cloud completion method, it would be unaffordable for our method to conduct comparisons on full completion benchmarks such as Completion3D~\citep{tchapmi2019topnet} or ShapeNet~\citep{chang2015shapenet} with thousands of point clouds.
% As it would be unaffordable for the test-time completion methods such as SDS-complete~\citep{kasten2024point} to conduct comparisons on full completion benchmarks such as Completion3D~\citep{tchapmi2019topnet} or ShapeNet~\citep{chang2015shapenet} with thousands of point clouds, we conduct comparisons on smaller test sets following SDS-complete~\citep{kasten2024point}.
Considering the impracticality of applying test-time completion methods~\citep{kasten2024point} to benchmarks like Completion3D~\citep{tchapmi2019topnet} or ShapeNet~\citep{chang2015shapenet} containing thousands of point clouds, We sampled an appropriate amount of test data following SDS-Complete~\citep{kasten2024point}.
% we follow SDS-complete~\citep{kasten2024point} to sample test data with comparable number as theirs.
% conduct comparisons on smaller representative test sets.
For synthetic data, we sample partial point clouds by sampling from various viewpoints around completely modeled objects from established sources~\citep{DBLP:conf/siggraph/KrishnamurthyL96,DBLP:journals/tog/DeCarloFRS03,DBLP:conf/siggraph/PraunFH00,DBLP:journals/tog/LipmanLC08}. 
For real scans, we use Redwood~\citep{choi2016large} following SDS-complete~\citep{kasten2024point}. Single scans are used as partial input while the ground truths are adopted by composing multiple scans. 
Comparisons on ShapeNet~\citep{chang2015shapenet} and Kitti~\citep{geiger2013vision} are presented in the appendix~\ref{sec:appendix}.
% We standardize all point clouds following PCN~\citep{yuan2018pcn}. 
% \textcolor{black}{As SDS-complete~\citep{kasten2024point} only provides evaluation dataset without codes,} 

% \textcolor{black}{As the only known zero-shot point cloud completion method, text prompt-based SDS-complete~\citep{kasten2024point} only provides evaluation dataset without codes. }
% In addition, it also requires extra manually defined text prompts for completion.
We compare our approach with state-of-the-art supervised methods including PointAttN\citep{wang2024pointattn}, PoinTr \citep{yu2021pointr}, SVDFormer \citep{Zhu_2023_ICCV}, AdaPoinTr \citep{yu2023adapointr}, \textcolor{black}{SeedFormer~\citep{zhou2022seedformer}, ShapeFormer~\citep{yan2022shapeformer}}. 
As SDS-complete~\citep{kasten2024point} only provide codes for the processing of Redwood dataset~\citep{choi2016large}, we implement corresponding comparisons on Redwood.
The evaluation metrics include the L1 Chamfer Distance (CD) and Earth Mover's Distance (EMD) \citep{fan2017point} that measure the similarity between the reconstructed point clouds and the ground truths. All metrics are multiplied with $10^2$ in subsequent comparisons.
We standardize point clouds and conduct comparisons at a resolution of 16,384 points following PCN \citep{yuan2018pcn}. Our results presented for comparisons on both synthetic data and real scans are averaged over three repeated experiments.
% All comparisons are performed at a resolution of 16,384 points following PCN \citep{yuan2018pcn}.
% To evaluate the performances of our method, we conduct experiments on objects from both synthetic data and real world scans. For the synthetic data, we sample partial point clouds from random views of collected complete objects~\citep{DBLP:conf/siggraph/KrishnamurthyL96,DBLP:journals/tog/DeCarloFRS03,DBLP:conf/siggraph/PraunFH00,DBLP:journals/tog/LipmanLC08}. 
% For the real world scans, we follow \citep{kasten2024point} for the settings of Redwood dataset~\citep{choi2016large}. Single scans are used as partial input, while the ground truths are adopted by composing multiple scans. 
% All point clouds are normalized to $-0.5 \sim 0.5$ following \citep{yuan2018pcn}.

% A few State-of-the-art methods PoinTr~\citep{yu2021pointr}, SnowflakeNet~\citep{xiang2022snowflake}, Seedformer~\citep{zhou2022seedformer}, and AdaPoinTr~\citep{yu2023adapointr} are adopted for comparison in this work.
% The performances of different methods are evaluated by Chamfer Distance (CD) and Earth Mover's Distance (EMD)~\citep{fan2017point} between completed point clouds and ground truths. All comparisons are conducted under the resolution of 16384 points following PCN~\citep{yuan2018pcn}.
\begin{figure*}[t]
	%%\vskip -0.3in
	\begin{center}
		\scalebox{1.0}{
			\centerline{\includegraphics[width=\linewidth]{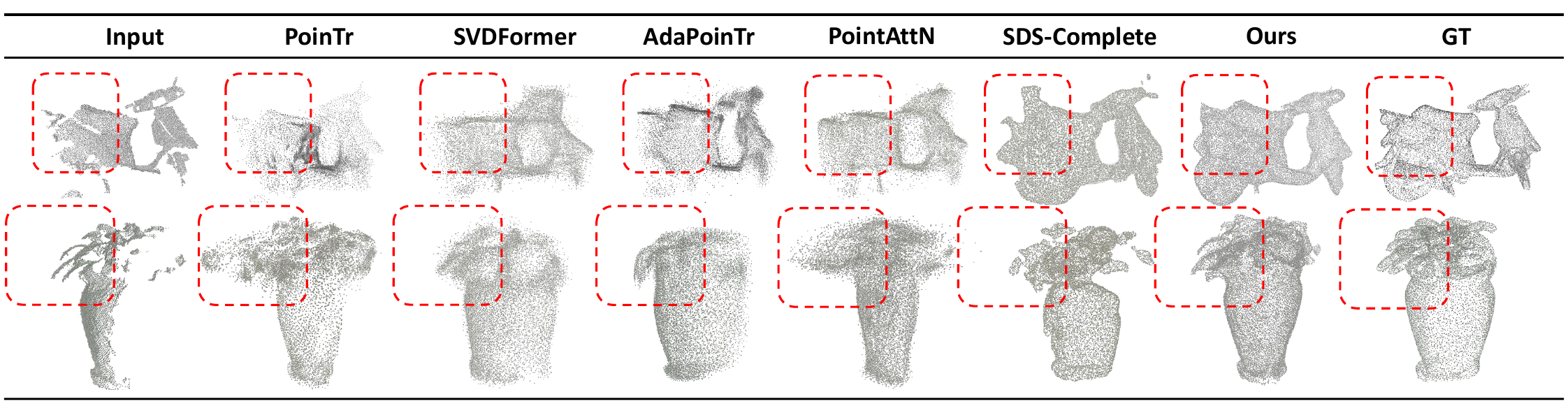}}
		}
		\vskip -0.1in
		\caption{Qualitative comparison on Redwood dataset~\citep{choi2016large,kasten2024point}.}
		\label{pic:quali_red}
	\end{center}
	\vskip -0.2in
\end{figure*}
\begin{table}[t]
		\normalsize
		\setlength\tabcolsep{1.2pt}
		\renewcommand\arraystretch{1.05}
  \caption{Quantitative comparison on Redwood dataset~\citep{choi2016large,kasten2024point}. For the convenience, we re-optimize and normalize the results of SDS-Complete consistently to $-0.5 \sim 0.5$.}
		\begin{center}
  % \vskip -0.1 in
			\scalebox{0.8}{
				\begin{tabular}{ccccccccccc}
\toprule
           & \multicolumn{5}{c}{In Domain}                             & \multicolumn{5}{c}{Out Domain}                             \\ \cmidrule{2-11}
Object   & Table     & Exe-Chair & Out-Chair & Old-Chair & Average   & Vase & Off Can   & Vespa     & Tricycle   & Average   \\
Metrics    & CD/EMD    & CD/EMD    & CD/EMD    & CD/EMD    & CD/EMD    & CD/EMD    & CD/EMD    & CD/EMD    & CD/EMD     & CD/EMD    \\
PoinTr     & 3.56/7.42 & 1.91/4.50 & \textbf{0.67}/1.41 & 2.48/6.28 & 2.16/4.90 & 3.84/6.55 & 3.76/5.83 & 1.84/3.94 & 5.91/11.63 & 3.84/6.99 \\
SeedFormer & 3.38/7.01 & 1.90/4.55 & 0.76/1.44 & 2.72/5.16 & 2.19/4.54 & 3.99/6.36 & 3.88/6.11 & 2.38/4.38 & 2.10/3.38  & 3.09/5.06 \\
% Snowflake  & 3.65/7.54 & 2.00/4.83 & 0.80/1.58 & 2.39/5.08 & 2.21/4.76 & 3.82/6.15 & 3.33/5.75 & 2.29/4.47 & \textbf{1.77}/3.50  & 2.80/4.96 \\
PointAttN  & 5.71/7.11 & 2.88/5.65 & 0.73/1.46 & 3.73/6.02 & 3.26/5.06 & 5.35/6.97 & 4.93/6.44 & 2.69/4.70 & \textbf{1.72}/3.59  & 3.67/5.43 \\
\textcolor{black}{ShapeFormer}  & 3.48/5.67 & 3.41/5.32 & 3.87/6.93 & 3.00/4.07 & 3.44/5.50 & 4.79/6.50 & 2.96/3.89 & 3.21/4.20 & 3.21/4.20  & 4.01/5.43 \\
% AnchorFormer  & 3.66/5.05 & 1.22/2.92 & 0.69/1.43 & 3.02/5.01 & 2.15/3.60 & 4.84/6.54 & 1.87/3.74 & 1.28/2.67 & 4.81/9.45  & 3.20/5.60 \\
SVDFormer  & 2.13/3.29 & 3.60/6.02 & 1.15/2.15 & 3.69/5.83 & 2.64/4.32 & 5.20/7.28 & 5.42/7.05 & 3.30/5.25 & 3.78/4.55  & 4.42/6.03 \\
AdaPoinTr  & 5.02/6.23 & 2.58/4.79 & 0.82/\textbf{1.38} & 3.62/5.61 & 3.01/4.50 & 5.14/6.48 & 4.47/6.32 & 1.94/3.52 & 1.83/3.67  & 3.34/4.98 \\
\midrule
SDS-Complete  & \textbf{1.35/2.30} & 1.96/2.65 & 2.51/3.92 & 2.77/3.77 & 2.15/3.16 & 3.00/5.25 & 3.79/4.28 & 3.36/5.73 & 3.18/3.49  & 3.33/4.69 \\
Ours       & 1.67/3.11 & \textbf{1.04/1.39} & 1.28/1.73 & \textbf{1.42/1.87} & \textbf{1.35/2.03} & \textbf{2.94/4.63} & \textbf{3.51/3.86} & \textbf{1.39/2.27} & 2.42/\textbf{1.94} & \textbf{2.57/3.17}  \\ \bottomrule
\end{tabular}
			}
		\end{center}
		 % \vskip -0.1in
		\label{quan_red}
		\vskip -0.2in
	\end{table}
\subsection{Comparison on Synthetic Point Clouds}
\label{sec:objs}
In this section, we conduct an evaluation on synthetic point clouds. The quantitative and qualitative results are presented in Table~\ref{quan_obj} and Fig.~\ref{pic:quali_obj}, respectively. 
\textcolor{black}{Existing network-based methods create noisy and incorrect shapes due to the discrepancies between their training data and the test data.}
% Our method significantly outperforms other methods on these data.
% \textcolor{black}{Existing methods }
As shown in Fig.~\ref{pic:quali_obj}, our method creates correct and reasonable completed results, which may benefit from abundant priors from the pre-trained diffusion model.
An interesting case is that our method completes an appropriate handle for the teapot in the first row of Fig.~\ref{pic:quali_obj} without any prompts and related geometries. It confirms that the pipeline can actually percept the actual categories of completed objects instead of simply inferring a shape to fill in the missing regions.
\begin{figure*}[t]
	%%\vskip -0.3in
	\begin{center}
		\scalebox{1.0}{
			\centerline{\includegraphics[width=\linewidth]{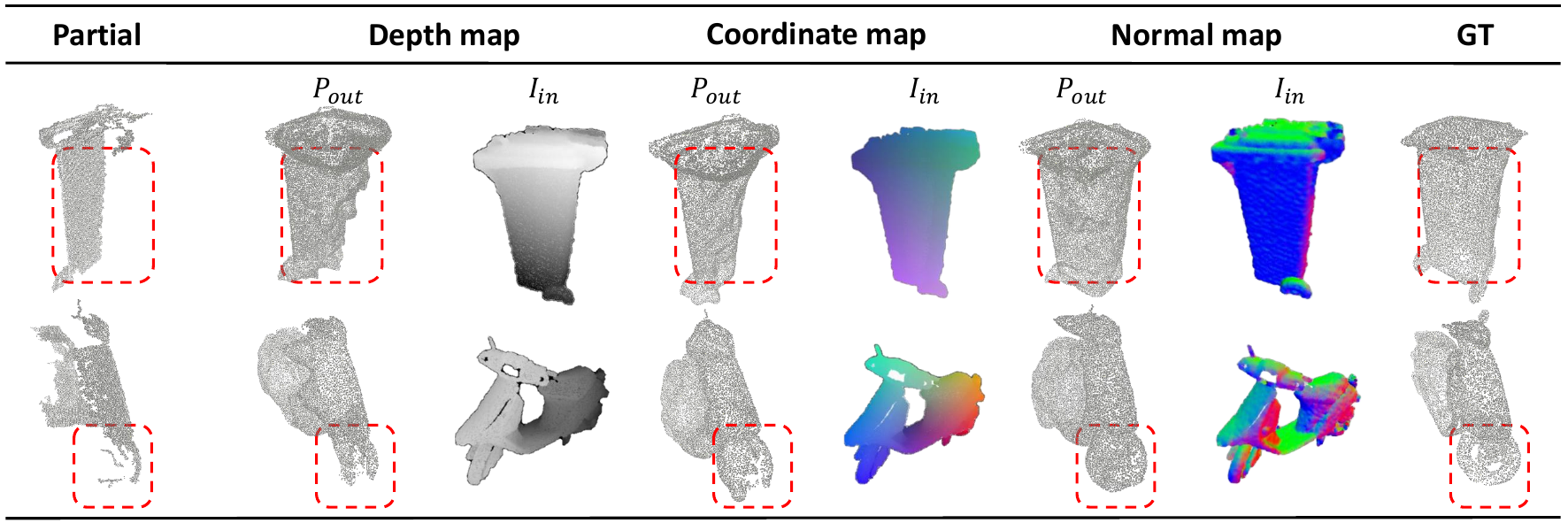}}
		}
		\vskip -0.1in
		\caption{ Qualitative comparison between different colorization strategies. $I_{in}$ and $P_{out}$ denote the colorized reference image and completed point clouds, respectively.}
		\label{pic:aba_pipe}
	\end{center}
	\vskip -0.25in
\end{figure*}

% \begin{table}[t]
% 		% \normalsize
% 		% \setlength\tabcolsep{1.2pt}
% 		% \renewcommand\arraystretch{1.05}
%   \caption{Quantitative comparison for different colorization strategies.}
% 		\begin{center}
%   % \vskip -0.1 in
% 			\scalebox{1.0}{
% 				\begin{tabular}{c|ccc}
% \toprule
%     & Depth & Coordinates & Normal(Ours)        \\ \midrule
% CD  & 1.60  & 1.44 & \textbf{1.38} \\ 
% EMD & 2.28  & 2.07 & \textbf{2.02} \\ \bottomrule
% \end{tabular}
% 			}
% 		\end{center}
% 		 % \vskip -0.1in
% 		\label{quan_red}
% 		\vskip -0.2in
% 	\end{table}

  \begin{table}[t]
    \centering
    \begin{minipage}{0.45\textwidth}
        \centering
        \normalsize
		\setlength\tabcolsep{1.5pt}
  		\caption{Ablation for colorization.}
    \vskip 0.1in
    \scalebox{1.0}{
        \begin{tabular}{c|ccc}
\toprule
    & Depth & Coordinates & Normal(Ours)        \\ \midrule
CD  & 2.25  & 2.01 & \textbf{1.96} \\ 
EMD &  2.88 & 2.64 & \textbf{2.60} \\ \bottomrule
\end{tabular}
                }
        \label{tab:aba_color}
	\vskip -0.2in
    \end{minipage}\hfill
    \begin{minipage}{0.45\textwidth}
        \centering
        \normalsize
		\setlength\tabcolsep{1.5pt}
  		\caption{Ablation for ZFC and PCE.}
    \vskip 0.1in
    \scalebox{0.9}{
%         \begin{tabular}{cccc|cc}
% \toprule
% Guidance & Pres & Surf & GP        & CD & EMD \\ \midrule
%     \checkmark     &      &      &  &  1.98  &   3.20  \\
%      \checkmark    &   \checkmark   &      &  &  1.96  &  3.07   \\
%     \checkmark     &   \checkmark   &    \checkmark  &           &  \textbf{1.50}  &   3.38  \\
%      \checkmark    &   \checkmark   &   \checkmark   &      \checkmark     &  1.96  &  \textbf{2.60}   \\ \bottomrule
% \end{tabular}
\begin{tabular}{cccc|cccc}
\toprule
\multirow{2}{*}{Guidance} & \multirow{2}{*}{Pres}     & \multirow{2}{*}{Surf}     & \multirow{2}{*}{GP}       & \multicolumn{2}{c}{Redwood}   & \multicolumn{2}{c}{Synthetic} \\
                          &                           &                           &                           & CD            & EMD           & CD            & EMD           \\ \midrule
\checkmark &                           &                           &                           & 1.98          & 3.20          & 3.35          & 6.01          \\
\checkmark & \checkmark &                           &                           & 1.97          & 3.07          & 2.55          & 4.41          \\
\checkmark & \checkmark & \checkmark &                           & \textbf{1.50} & 3.38          & \textbf{1.17} & 3.74          \\
\checkmark & \checkmark & \checkmark & \checkmark & 1.96          & \textbf{2.60} & 1.43          & \textbf{1.88} \\ \bottomrule
\end{tabular}
                }
        \label{tab:aba_com}
        % \vskip -0.2in
    \end{minipage}
    \vskip -0.2in
\end{table}

\begin{figure*}[t]
	%%\vskip -0.3in
	\begin{center}
		\scalebox{1.0}{
			\centerline{\includegraphics[width=\linewidth]{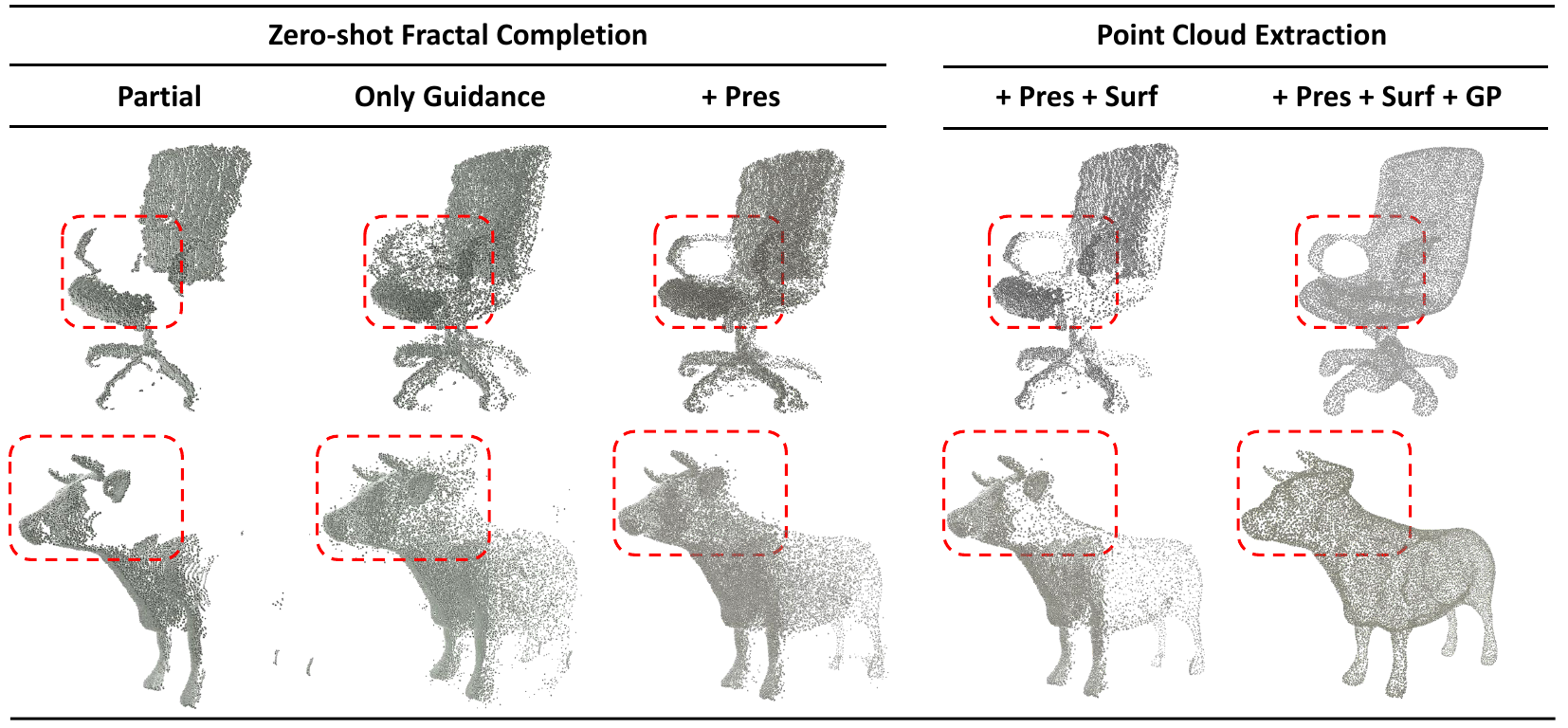}}
		}
		\vskip -0.1in
		\caption{Qualitative ablation study for ZFC and PCE. Surf, Pres, and GP denote Gaussian Surface Extraction, Preservation Constraints, and Grid Pulling, respectively.}
		\label{pic:aba_zfc}
	\end{center}
	\vskip -0.2in
\end{figure*}

% \begin{table}[t]
% 		% \normalsize
% 		% \setlength\tabcolsep{1.2pt}
% 		% \renewcommand\arraystretch{1.05}
%   \caption{Quantitative comparison for different colorization strategies.}
% 		\begin{center}
%   % \vskip -0.1 in
% 			\scalebox{1.0}{
% \begin{tabular}{cccc|cc}
% \toprule
% Guidance & Pres & Surf & GP        & CD & EMD \\ \midrule
%          &   \checkmark   &      &  &    &     \\
%          &   \checkmark   &   \checkmark   &  &    &     \\
%          &   \checkmark   &    \checkmark  &     \checkmark      &    &     \\
%          &   \checkmark   &   \checkmark   &      \checkmark     &    &     \\ \bottomrule
% \end{tabular}
% 			}
% 		\end{center}
% 		 % \vskip -0.1in
% 		\label{quan_red}
% 		\vskip -0.2in
% 	\end{table}
 
\subsection{Comparison on Real Scans}
\label{sec:redwood}
We follow SDS-complete~\citep{kasten2024point} for the comparison on real scans from Redwood~\citep{choi2016large}. 
% Following SDS-complete~\citep{kasten2024point}, s
Scans are divided into the "in domain" categories similar as training datasets of existing completion networks~\citep{yu2021pointr,zhou2022seedformer,yu2023adapointr,wang2024pointattn}, and "out domain" categories unseen during their training. 
The qualitative and quantitative comparison results are illustrated in Fig.~\ref{pic:quali_red} and Table~\ref{quan_red}, respectively.
As shown in Table~\ref{quan_red}, our method outperforms other methods on both "in domain" and "out domain" models, which further confirms the effectiveness and generalizability of our method.
\textcolor{black}{Existing fully-supervised methods may perform inferior even on the in-domain objects as illustrated in Table~\ref{quan_red}, which reveals their limitation on datasets differing from the training one. By introducing abundant priors from 2D diffusion model~\citep{liu2023zero}, our method can achieve robust completion for objects across different datasets.}

% We can see that defects show up in circled regions after removing t. The reason may be that the original depth images lack of textures, where the colorization with pre-trained depth-conditioned ControlNet adds clearer textures to the initial view. It is beneficial for subsequent view dependent guidance in ZFC.
% Please check the supplementary for more details about the colorization.
% As for the Grid Pulling module, it obviously generate quite uniform point clouds from the quite non-uniform completed results from ZFC.
\subsection{Ablation Study for Colorization Strategies in PGI}
% To validate the effect of colorization in the whole completion framework, we conduct an ablation study by removing the colorization from the framework. Removing the colorization means that we directly use the depth image $D_{in}$ as the reference image $I_{in}$ without colorization from depth-conditioned ControlNet~\citep{zhang2023adding}.
% \textcolor{black}{To verify the necessity of using normal vectors for colorization, we compare the completion performances with other straightforward colorization strategies, including depth values and normalized coordinates.
% As presented in Fig.~\ref{pic:aba_pipe}, other colorization strategies perform obviously inferior to the normals vectors in the circled regions.
% The reason may be that normal vectors can help reflect changes of the surfaces into the colors more obviously, which can better show the geometrical characteristics in the reference image. }
% As discussed in Sec.~\ref{sec:metho}, we set the color attributes in Partial Gaussian Initialization as scaled normal vectors, which 
To confirm the necessity of using normal map for colorization in Partial Gaussian Initialization, we compare their performances against other strategies including using depth values and normalized coordinates. 
As shown in Fig.~\ref{pic:aba_pipe}, these alternative strategies are clearly outperformed by the normal map composed of normal vectors, particularly in the circled areas. This superiority likely stems from the ability of normal vectors to more distinctly reflect surface changes in colors, thus better capturing the geometric characteristics in the reference image.
We also provide quantitative comparisons of different colorization strategies in Table~\ref{tab:aba_color}, using average metrics from in-domain and out-of-domain Redwood dataset. The results show that the normal map consistently outperforms other methods.
\subsection{Ablation Study for ZFC and PCE}
% \noindent\textbf{Abation for components in ZFC.} 
% Zero-shot Fractal Completion consists of view dependent guidance, Gaussian surface extraction, Preservation Constraint, and Grid Pulling module. 
In this work, we propose ZFC to introduce diffusion priors to infer the missing regions, and PCE to extract uniform point clouds from the 3D Gaussian centers.
ZFC is composed of view dependent guidance and Preservation Constraint, while PCE consists of Gaussian surface extraction and Grid Pulling.
From Fig.~\ref{pic:aba_zfc}, we can see that our method with all components have uniform and reasonable completed results.
In PCE, GP obviously generates quite uniform point clouds from the non-uniform ones directly acquired from 3D Gaussians. Gaussian Surface Extraction operation extracts the surface from relatively disorganized Gaussian centers.
%Removing the Preservation Constraint will produce redundant geometrical structures shown in the \textcolor{black}{circled} regions of Fig.~\ref{pic:aba_zfc}. 
In ZFC, view dependent guidance creates coarse results with relatively correct overall shapes.
Preservation Constraint avoids redundant shapes by introducing strict constraints between partially observed points and existing partial point clouds.

We also provide quantitative ablation study for our proposed components in Table~\ref{tab:aba_com}. 
We evaluated our method on both Redwood and synthetic datasets. The results demonstrate that the Preservation Constraint improves performance compared to standard view-dependent diffusion guidance. Although Gaussian surface extraction significantly enhances the CD metric by selecting surface points, it negatively affects the EMD metric due to the high non-uniformity, as shown in the fourth column of Fig.~\ref{pic:aba_zfc}. In contrast, the final Grid Pulling (GP) module acquire more uniform surface points, leading to better EMD performance, although the CD metric experiences a slight decline due to precision loss caused by potential deformations in GP. More detailed ablation study can be found in the appendix~\ref{sec:appendix}.
% We can explore to reduce the deformation in GP in our future work.
% We can further explore better solutions for this problem.
%geometrical structures 
% We can also see that the Gaussian Surface Extraction operation extracts the surface from relatively disorganized Gaussian centers.

\section{Limitation}
\textcolor{black}{Our method shares similar limitations as claimed by SDS-complete~\citep{kasten2024point}.
%It is slower than network-based methods as a test-time completion method.
As a test-time completion method, although our method does not require any training, the optimization on the test data would take relatively long time cost.
For instance, completing a point cloud from the Redwood dataset takes approximately 15 minutes with our method on a RTX A6000 GPU.
However, our framework is much more efficient than the existing test-time method SDS-complete~\citep{kasten2024point}, which takes up to 1950 minutes for optimization as reported in their supplementary material. 
% compared to the average of 1900 minutes reported by SDS-complete~\citep{kasten2024point} in their supplementary material. 
Please check the appendix~\ref{sec:failure} for failure cases of our method and additional implementation details.}

\section{Conclusion}
In this work, we introduce a test-time point cloud completion framework that leverages the rich priors from 2D diffusion models~\citep{liu2023zero,zhang2023adding} through 3D Gaussian Splatting rendering, which can robustly complete collected partial 3D point clouds without any requirements of training.
Our framework consists of three main components: Partial Gaussian Initialization (PGI), Zero-shot Fractal Completion (ZFC), and Point Cloud Extraction (PCE).
In PGI, we initialize 3D Gaussians using the partial point cloud to render a reference image from an estimated reference viewpoint. 
We then employ ZFC to infer the missing regions of the partial point cloud by optimizing 3D Gaussians, using view-dependent guidance conditioned on the reference image. 
Finally, with PCE, we extract uniformly completed point clouds from 3D Gaussian centers.
Our method outperforms both existing network-based and test-time approaches in achieving robust completion across multiple categories of both synthetic and real scanned data.



\bibliography{egbib}
\bibliographystyle{iclr2025_conference}

\appendix
\section{Appendix}
\input{X_suppl}

\end{document}

%% file: math_commands.tex
\usepackage{amsmath,amsfonts,bm}

\def\eqref#1{equation~\ref{#1}}
\def\1{\bm{1}}

\DeclareMathAlphabet{\mathsfit}{\encodingdefault}{\sfdefault}{m}{sl}
\SetMathAlphabet{\mathsfit}{bold}{\encodingdefault}{\sfdefault}{bx}{n}

%% file: X_suppl.tex
% \clearpage
% \setcounter{page}{1}
% \maketitlesupplementary

% \section{Rationale}
% \label{sec:rationale}
% % 
% Having the supplementary compiled together with the main paper means that:
% % 
% \begin{itemize}
% \item The supplementary can back-reference sections of the main paper, for example, we can refer to \cref{sec:intro};
% \item The main paper can forward reference sub-sections within the supplementary explicitly (e.g. referring to a particular experiment); 
% \item When submitted to arXiv, the supplementary will already included at the end of the paper.
% \end{itemize}
% % 
% To split the supplementary pages from the main paper, you can use \href{https://support.apple.com/en-ca/guide/preview/prvw11793/mac#:~:text=Delete%20a%20page%20from%20a,or%20choose%20Edit%20%3E%20Delete).}{Preview (on macOS)}, \href{https://www.adobe.com/acrobat/how-to/delete-pages-from-pdf.html#:~:text=Choose%20%E2%80%9CTools%E2%80%9D%20%3E%20%E2%80%9COrganize,or%20pages%20from%20the%20file.}{Adobe Acrobat} (on all OSs), as well as \href{https://superuser.com/questions/517986/is-it-possible-to-delete-some-pages-of-a-pdf-document}{command line tools}.
% \section{Supplementary}
\label{sec:appendix}
\subsection{Details of Parameter Settings}
\label{sec:hyperpara}
In Table~\ref{tab_para}, we provide detailed information on the hyper-parameters discussed in Sec.~\ref{sec:metho}. Our experiments are conducted on RTX A6000/A5000 GPU, with PyTorch 1.12 and CUDA 11.6.
% Despite the apparent presence of multiple parameters, \emph{the settings remain robust for both datasets utilized.}
	\begin{table}[ht]
 \vskip -0.1in
		\normalsize
		\setlength\tabcolsep{2.0pt}
  	\caption{The setting of mentioned hyper-parameters in Sec.~\ref{sec:metho}.}
		\begin{center}
  % \vskip -0.2in
			\scalebox{0.9}{
				\begin{tabular}{cc}
\toprule
            & Hyper-parameters                                \\ \cmidrule{2-2}
$w_0\sim w_3$       & 1e-3, 1e3, 1e2, 0.1 \\
$\delta, \sigma_0, \sigma_n$    & 0.01, 0.005, 0.05                                \\
Iterations & 1000 (ZFC),  5000 (PCE)                                     \\ \bottomrule
% $n$           & 5                                        \\ \hline
\end{tabular}
			}
		\end{center}
		% \vskip -0.2in
		\label{tab_para}
		\vskip -0.1in
	\end{table}

\begin{figure}[t]
	\vskip -0.1in
	\begin{center}
		\scalebox{0.7}{
			\centerline{\includegraphics[width=\linewidth]{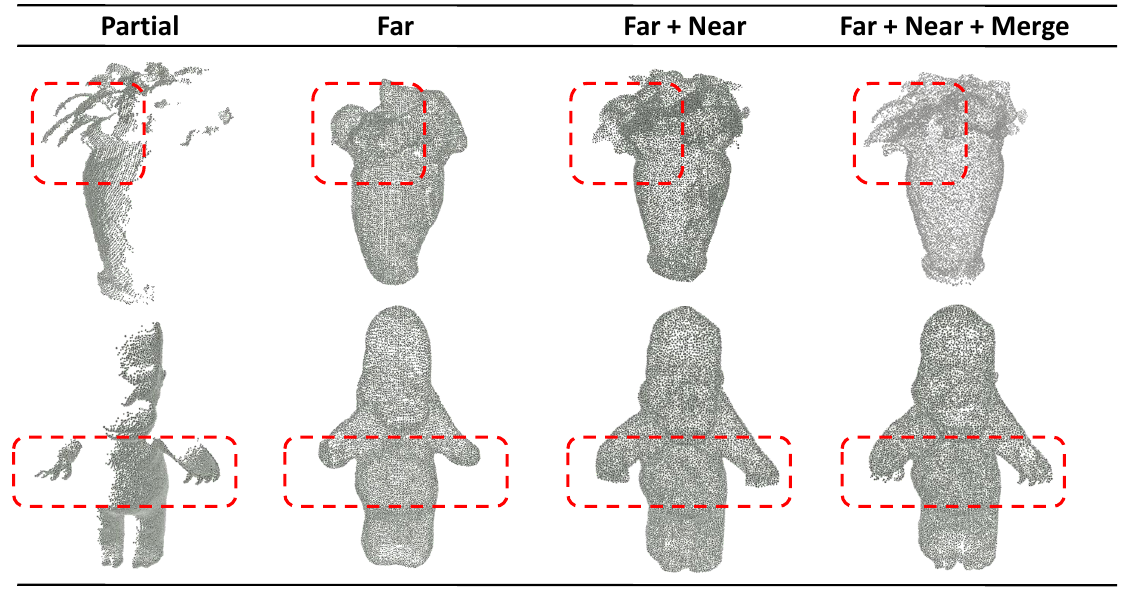}}
		}
		\vskip -0.1in
		\caption{Ablation study for Grid Pulling module. Far, Near, and Merge denote the $L_{far}$, $L_{near}$, and merge layer $g_m(\cdot)$, respectively.}
		\label{pic:aba_gp}
	\end{center}
	\vskip -0.2in
\end{figure}

% \vspace{2mm}
\noindent \textbf{Ablation for Grid Pulling.}
Grid Pulling (GP) module is proposed to resample uniform and regular point clouds from non-uniform $P_{surf}$ in the Point Cloud Extraction. 
As claimed in Sec.~\ref{sec:gp}, $L_{far}$ and $L_{near}$ are used to optimize an continuous surface presented by MLP while Merge layer is introduced to merge output point clouds with partial input. 
From results in Fig.~\ref{pic:aba_gp}, we can observe that $L_{far}$ and $L_{near}$ contribute to overall contours and local shapes, respectively. Nonetheless, they are still limited to the over-smoothed results.
The merge layer helps preserve local geometrical details in the circled regions from the partial input point cloud.
We also provide a quantitative comparison on GP module in Table~\ref{tab:aba_gp}. We can see that each component in GP contributes to the final performance.
\begin{table}[t]
		% \normalsize
		% \setlength\tabcolsep{1.2pt}
		% \renewcommand\arraystretch{1.05}
  \caption{Quantitative comparison for Grid Pulling module evaluated on Redwood dataset.}
		\begin{center}
  % \vskip -0.1 in
			\scalebox{1.0}{
				\begin{tabular}{c|ccc}
\toprule
    & No Merge \& Near & No Merge & Ours      \\ \midrule
CD  &      3.11        &    2.04     & \textbf{1.96}  \\
EMD &      3.11        &    2.64     & \textbf{2.60} \\ \bottomrule
\end{tabular}
			}
		\end{center}
		 % \vskip -0.1in
		\label{tab:aba_gp}
		\vskip -0.2in
	\end{table}
\begin{figure*}[ht]
	%%\vskip -0.3in
	\begin{center}
		\scalebox{0.7}{
			\centerline{\includegraphics[width=\linewidth]{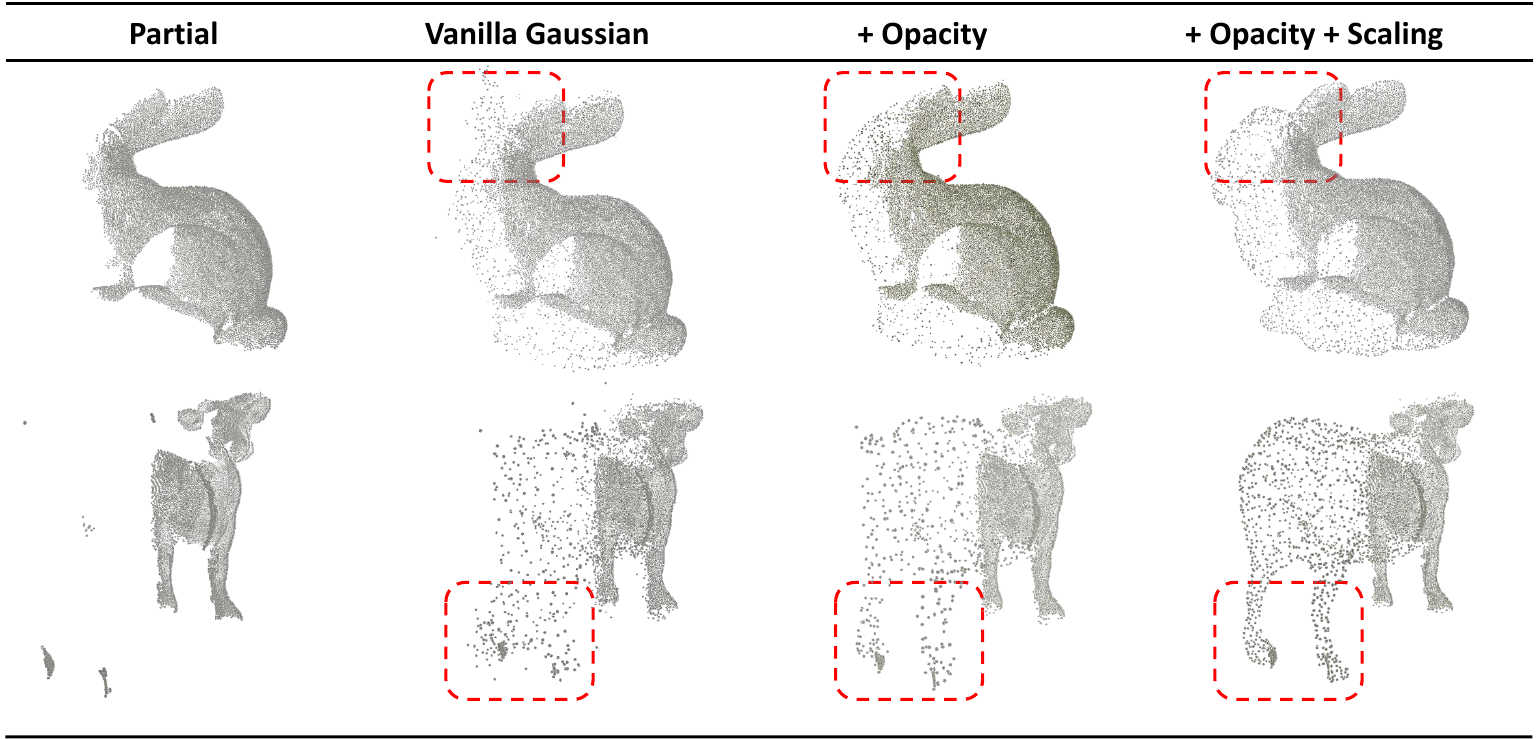}}
		}
		\vskip -0.1in
		\caption{Ablation Study for the 3D Gaussian modifications. Scaling and Opacity denotes the parameter-shared scalar scaling and binary opacity operations mentioned in Sec.~\ref{sec:zfc}, respectively.}
		\label{pic:aba_gaussian}
	\end{center}
	\vskip -0.3in
\end{figure*}

\begin{table}[ht]
		% \normalsize
		% \setlength\tabcolsep{1.2pt}
		% \renewcommand\arraystretch{1.05}
  \caption{Quantitative comparison for 3D Gaussian modifications evaluated on Redwood dataset.}
		\begin{center}
  % \vskip -0.1 in
			\scalebox{1.0}{
				\begin{tabular}{c|ccc}
\toprule
    & No Opacity \& Scaling & No Scaling & Ours      \\ \midrule
CD  &       6.35       &    2.50     &  \textbf{1.96}  \\
EMD &       10.48       &    3.58     &  \textbf{2.60} \\ \bottomrule
\end{tabular}
			}
		\end{center}
		 % \vskip -0.1in
		\label{tab:aba_gaus}
		\vskip -0.2in
	\end{table}
 
\begin{figure*}[ht]
	%%\vskip -0.3in
	\begin{center}
		\scalebox{0.7}{
			\centerline{\includegraphics[width=\linewidth]{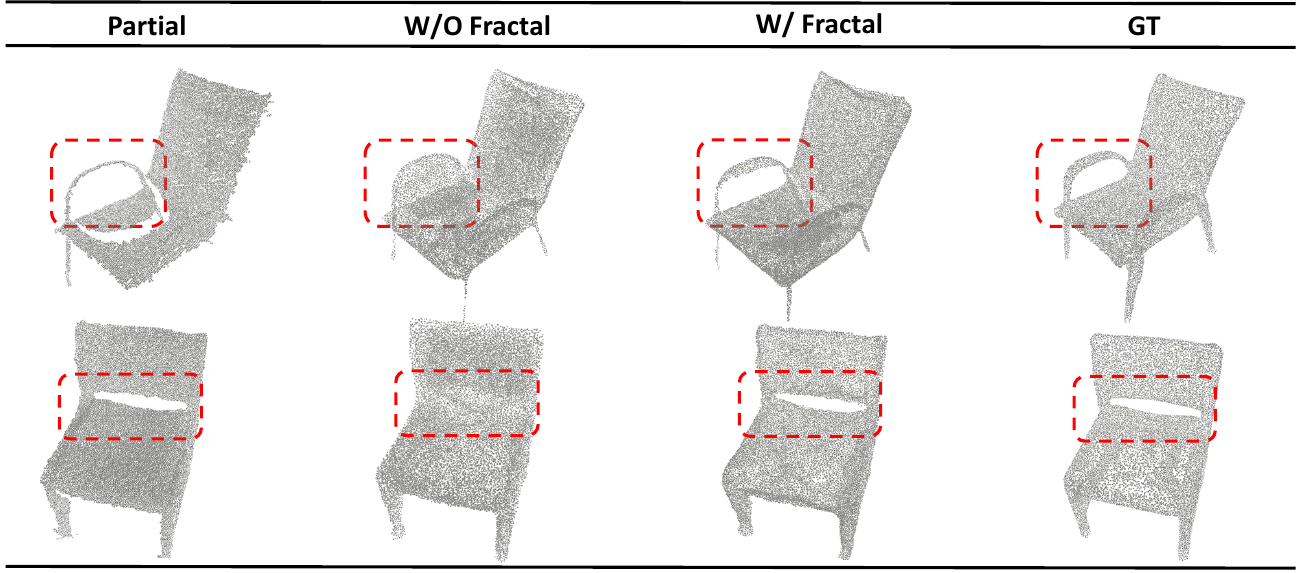}}
		}
		\vskip -0.1in
		\caption{Ablation Study for the Fractal completion strategy. W/ Fractal and W/O Fractal denote using and not using Fractal completion strategy, respectively.}
		\label{pic:aba_fractal}
	\end{center}
	\vskip -0.2in
\end{figure*}
\begin{table}[ht]
  \caption{Quantitative comparison for the Fractal strategy evaluated on Redwood dataset.}
		\begin{center}
  % \vskip -0.1 in
			\scalebox{1.0}{
				\begin{tabular}{c|cc}
\toprule
    & w/o Frac & w/ Frac      \\ \midrule
CD  &      2.00        &    \textbf{1.96}       \\
EMD &      2.69        &     \textbf{2.60}      \\ \bottomrule
\end{tabular}
			}
		\end{center}
		 \vskip -0.1in
		\label{tab:aba_frac}
		\vskip -0.2in
	\end{table}

 \begin{figure*}[ht]
	%%\vskip -0.3in
	\begin{center}
		\scalebox{0.6}{
			\centerline{\includegraphics[width=\linewidth]{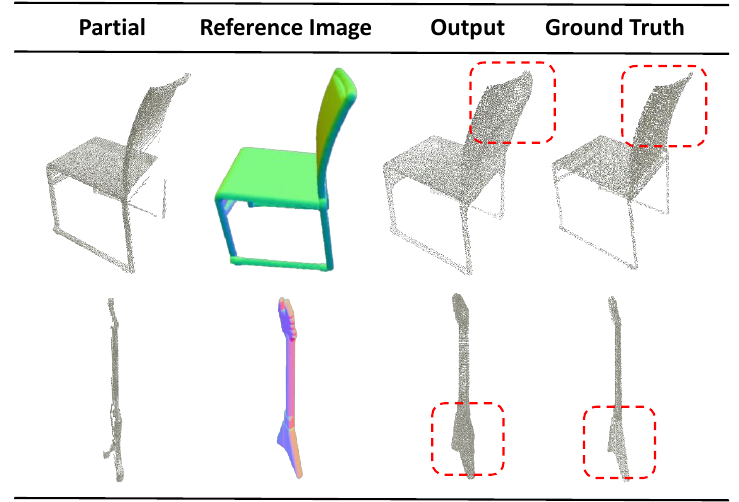}}
		}
		\vskip -0.1in
		\caption{Some failure cases.}
		\label{pic:failures}
	\end{center}
	\vskip -0.3in
\end{figure*}

%\subsection{Ablation Study}
%\noindent \textbf{Ablation for pipeline.} 
\subsection{Ablation Study for Modifications of 3D Gaussians}
As presented in Sec.~\ref{sec:zfc}, we make a few modifications to the original 3D Gaussian Splatting~\citep{kerbl20233d} including the parameter-shared scalar scaling definition and binary opacity estimation. In this section, we conduct a few experiments to validate the effectiveness of these proposed operations. 
To better illustrate their performances, we conduct comparisons based on $P_{surf}$ directly acquired from the Gaussian centers in ZFC.
The results are presented in Fig.~\ref{pic:aba_gaussian}. 

Adopting a shared scalar scaling helps in revealing more defined geometric details in the point cloud completion task.
The original settings of separate scaling across different 3D Gaussians tend to produce blurring edges and lose finer details.
% The , which allowed for rotation and variable scaling factors across different 3D Gaussians, tends to blur edge definitions and obscure finer details.
In addition, the binary opacity operation obviously reduce the noises in $P_{surf}$. With the original opacity settings, a considerable number of 3D Gaussians with moderate opacity values would scatter around the actual surfaces, blurring the distinction between the object and its surroundings. 
The binary opacity method effectively eliminates this issue, ensuring a cleaner bounding and more accurate surface representation.
As shown in Table~\ref{tab:aba_gaus}, the modifications on 3D Gaussians have significant influence on the completion performances.
% The separated and rotatable scaling in original Gaussian Splatting 

% Besides, as our method introduces diffusion priors 
% These problems may be resolved by the combination of 3D supervision and 2D diffusion priors, which means to expand to 

\subsection{Effect of the Fractal Completion Strategy}
As illustrated in Fig.~\ref{pic:framework}, we introduce the fractal completion strategy in ZFC by optimizing 3D Gaussians $G_m$ together with frozen 3D Gaussians $G_{in}$ initialized from partial point clouds. In this section, we conduct experiments to verify the effect of this strategy. A few visualized examples are presented in Fig.~\ref{pic:aba_fractal}. 
When not using fractal completion strategy, we directly optimize 3D Gaussians $G_m$ for all structures without concatenation with $G_{in}$.
We observe that completions without the fractal completion strategy tend to overlook some shape details present in the input partial point clouds. The quantitative results in Table~\ref{tab:aba_frac} further validate the advantages of the fractal strategy.
% fixing the 3D Gaussian parameters initialized from partial point clouds and optimizing 

\subsection{Failure Cases}
\label{sec:failure}
Fig.~\ref{pic:failures} presents some failure cases. Our method encounters similar problems as SDS-complete~\citep{kasten2024point} when generating thin surfaces in occluded areas. 
In these cases, 2D diffusion priors tend to imagine the thin occluded regions as reasonable but thicker structures as shown in Fig.~\ref{pic:failures}.
% Introducing constraints from pre-trained 3D networks could potentially address this problem by introducing a bit 3D prior.
% Introducing some 3D priors for constraints during completion could potentially address this problem.
This problem could be potentially addressed by fine-tuning the 2D diffusion priors, or introducing some regularization during the optimization process.
We will explore it in our future work.

\subsection{Discussion about the Incompleteness of Reference Observation}
% \subsection{What if the Reference Observation is Incomplete?}
As discussed in Sec.~\ref{sec:zfc} and Sec.~\ref{sec:assumption}, we utilize the reference image $I_{in}$ as a guiding condition for completion using the diffusion model~\citep{liu2023zero} on the point clouds.
The reference image is generated from an estimated viewpoint aimed at capturing the most complete observation of the point cloud $P_{in}$. However, due to sensor limitations, observations from this viewpoint may still exhibit some degree of incompleteness in certain cases.
In this section, we delve into a brief discussion regarding the impact of such incompleteness.

\begin{figure*}[ht]
	% \vskip -0.3in
	\begin{center}
		\scalebox{1.0}{
			\centerline{\includegraphics[width=\linewidth]{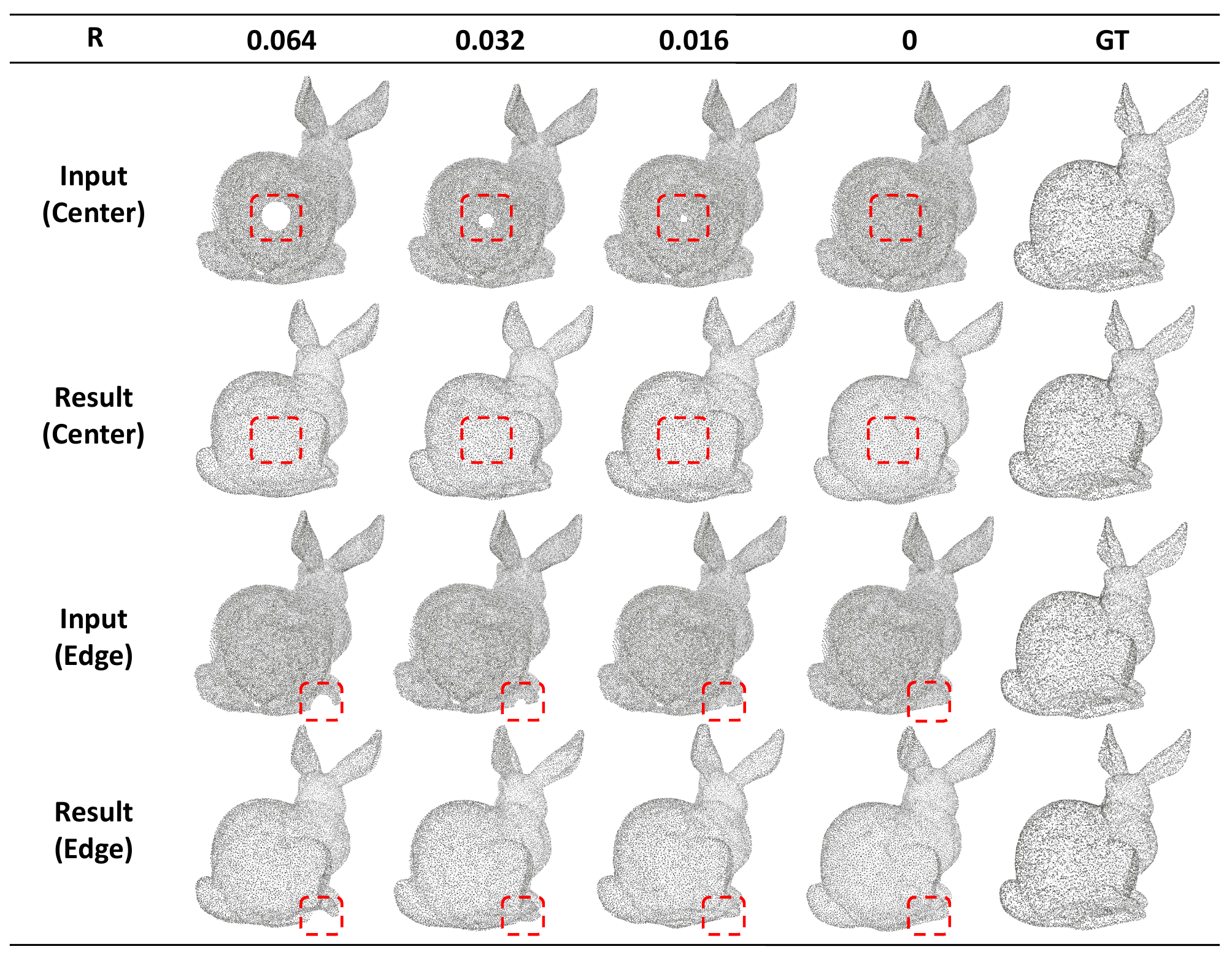}}
		}
		\vskip -0.1in
		\caption{Discussion about the incompleteness of reference observation. Center and Edge denote the locations of different missing regions. \textbf{$R$} is the radii of eliminated regions.}
		\label{pic:incom}
	\end{center}
	% \vskip -0.3in
\end{figure*}

As illustrated in Fig.~\ref{pic:incom}, we eliminate points within varying-sized regions located at the center and edges of the partial point cloud observed from the reference viewpoint mentioned in Sec.~\ref{sec:zfc}. 
It is evident that our approach effectively fills in the missing regions at the center of the point clouds, demonstrating its capability to infer missing areas based on surrounding points  to a certain extent.
Moreover, our method successfully addresses small gaps in the edge regions, as depicted in the second and third columns of Fig.~\ref{pic:incom}. Although large gaps in the edge regions may result in defects within the respective areas, they do not impede completion in other regions.
% Besides, our method can also complete small missing in the edge regions.
% Therefore, the completion process may be 
% Zero-shot Fractal Completion (ZFC) 

\begin{figure*}[t]
	%%\vskip -0.3in
	\begin{center}
		\scalebox{1.0}{
			\centerline{\includegraphics[width=\linewidth]{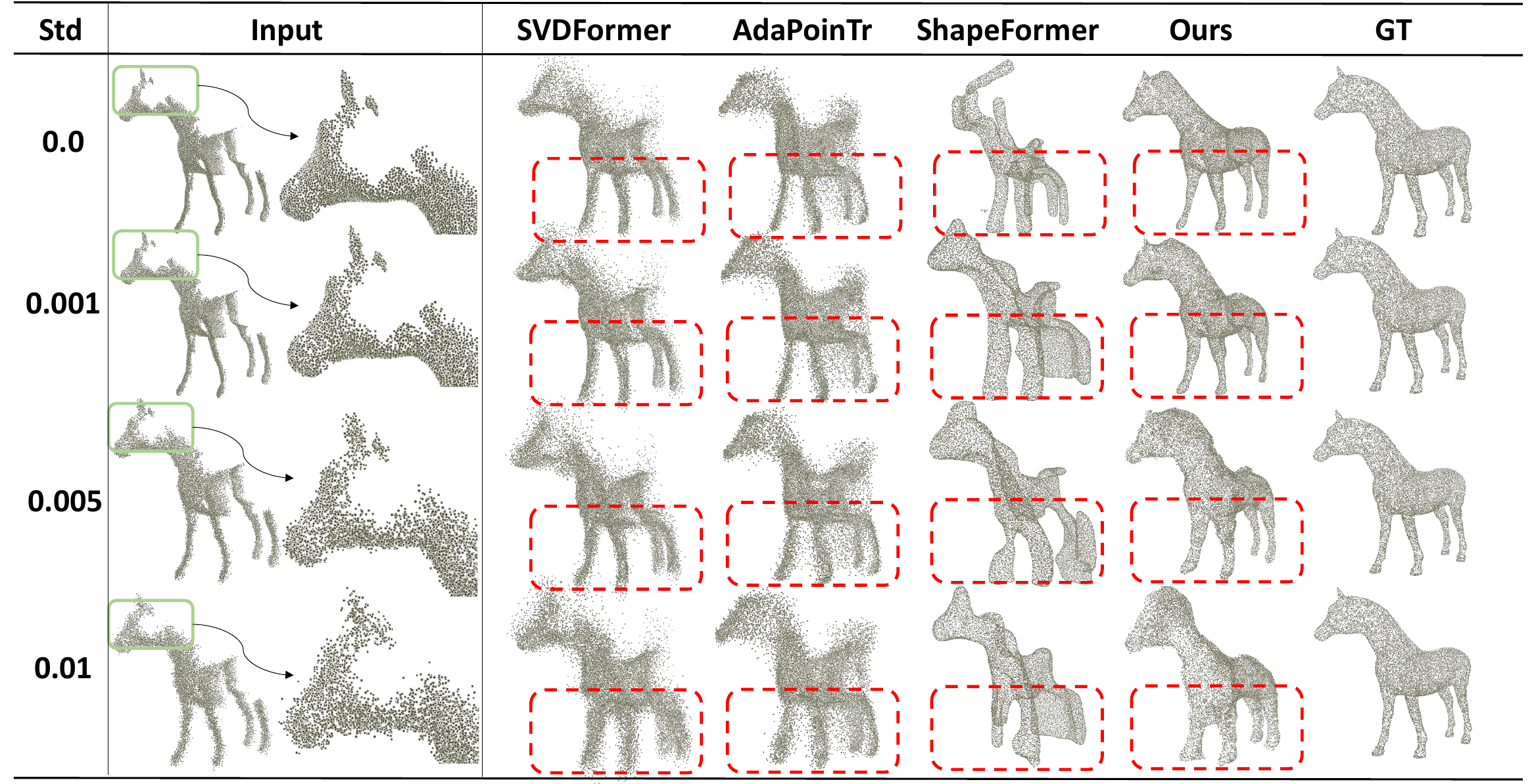}}
		}
		\vskip -0.1in
	\caption{\textcolor{black}{Qualitative comparisons under different noise perturbations. Std denotes the Standard deviation of added noises. The \textcolor{green}{green} box marks a local area of a noised point cloud.}}
	\label{pic:noise}
	\end{center}
	\vskip -0.2in
\end{figure*}

\textcolor{black}{In addition to the incompleteness of observations, partial point clouds may also be affected by noise due to poor natural illumination or reflections from object surfaces. To evaluate the robustness of our method to such noise, we introduce varying levels of noise to the synthetic partial point clouds described in Sec.\ref{sec:datasets}. The quantitative results are summarized in Table\ref{tab:noises}, and qualitative comparisons are shown in Fig.~\ref{pic:noise}.
While the performance of our method decreases as the noise level increases, it consistently outperforms existing approaches. As illustrated in Fig.~\ref{pic:noise}, noise with a standard deviation of 0.01 introduces noticeable blurring to the input partial points, yet our method is still able to recover the overall contour effectively. This demonstrates that our approach exhibits a degree of robustness to noise.
The primary contribution of this work lies in the development of a practical framework that leverages 2D diffusion priors for 3D point cloud completion. Comparisons on real scans from the Redwood dataset~\cite{choi2016large} validate the effectiveness of our method in handling real-world data. Enhancing its robustness may further remain a promising direction for future research.
}

\begin{table}[ht]
 % \vskip -0.2in
  \caption{\textcolor{black}{Quantitative comparisons on noised input point clouds. Std denotes the Standard deviation of added noises.}}
		\begin{center}
  \setlength\tabcolsep{1.2pt}
			\scalebox{1.0}{
				\begin{tabular}{c|ccccccc}
\toprule
        & PoinTr    & Seedformer & PointAttN & SVDFormer & ShapeFormer & AdaPoinTr & Ours               \\
Std & CD/EMD    & CD/EMD     & CD/EMD    & CD/EMD    & CD/EMD      & CD/EMD    & CD/EMD             \\ \midrule
0       & 4.10/5.63 & 4.21/5.93  & 5.72/6.54 & 4.79/5.29 & 3.68/4.44   & 5.59/6.04 & \textbf{1.43/1.88} \\
0.001   & 4.15/5.59 & 4.17/5.91  & 5.76/6.55 & 4.73/5.22 & 3.65/4.52   & 5.59/6.04 & \textbf{1.53/1.87} \\
0.005   & 4.24/5.83 & 4.31/6.11  & 5.75/6.59 & 4.92/5.52 & 4.03/4.89   & 5.65/6.19 & \textbf{2.02/2.25} \\
0.01    & 4.16/5.85 & 4.34/6.22  & 5.62/6.73 & 4.86/5.75 & 4.06/5.00   & 5.57/6.44 & \textbf{3.18/4.07} \\ \bottomrule
\end{tabular}
			}
		\end{center}
		 % \vskip -0.1in
		\label{tab:noises}
		\vskip -0.1in
	\end{table}
    \begin{table}[t]
  \caption{\textcolor{black}{Quantitative comparisons under different incompleteness levels. The levels denote how many depth maps are used to construct the partial input.}}
		\begin{center}
  % \vskip -0.1 in
  \setlength\tabcolsep{1.2pt}
			\scalebox{1.0}{
				\begin{tabular}{c|ccccccc}
\toprule
      & PoinTr    & Seedformer & PointAttN & SVDFormer & ShapeFormer & AdaPoinTr & Ours               \\
Level & CD/EMD    & CD/EMD     & CD/EMD    & CD/EMD    & CD/EMD      & CD/EMD    & CD/EMD             \\ \midrule
1     & 3.77/5.13 & 4.16/6.02  & 5.52/6.29 & 4.63/5.08 & 3.30/4.07   & 5.33/5.82 & \textbf{1.86/2.01} \\
3     & 3.61/5.10 & 3.92/5.93  & 5.45/6.28 & 4.36/5.02 & 3.42/4.26   & 5.28/5.82 & \textbf{1.76/2.04} \\
7     & 3.06/4.95 & 3.48/5.72  & 5.18/6.13 & 4.16/4.95 & 3.00/3.77   & 5.26/5.85 & \textbf{1.48/1.87} \\ \bottomrule
\end{tabular}
			}
		\end{center}
		 % \vskip -0.1in
		\label{tab:levels}
		\vskip -0.2in
	\end{table}

\subsection{Discussion about different Incompleteness levels}
\begin{figure*}[t]
	%%\vskip -0.3in
	\begin{center}
		\scalebox{0.9}{
			\centerline{\includegraphics[width=\linewidth]{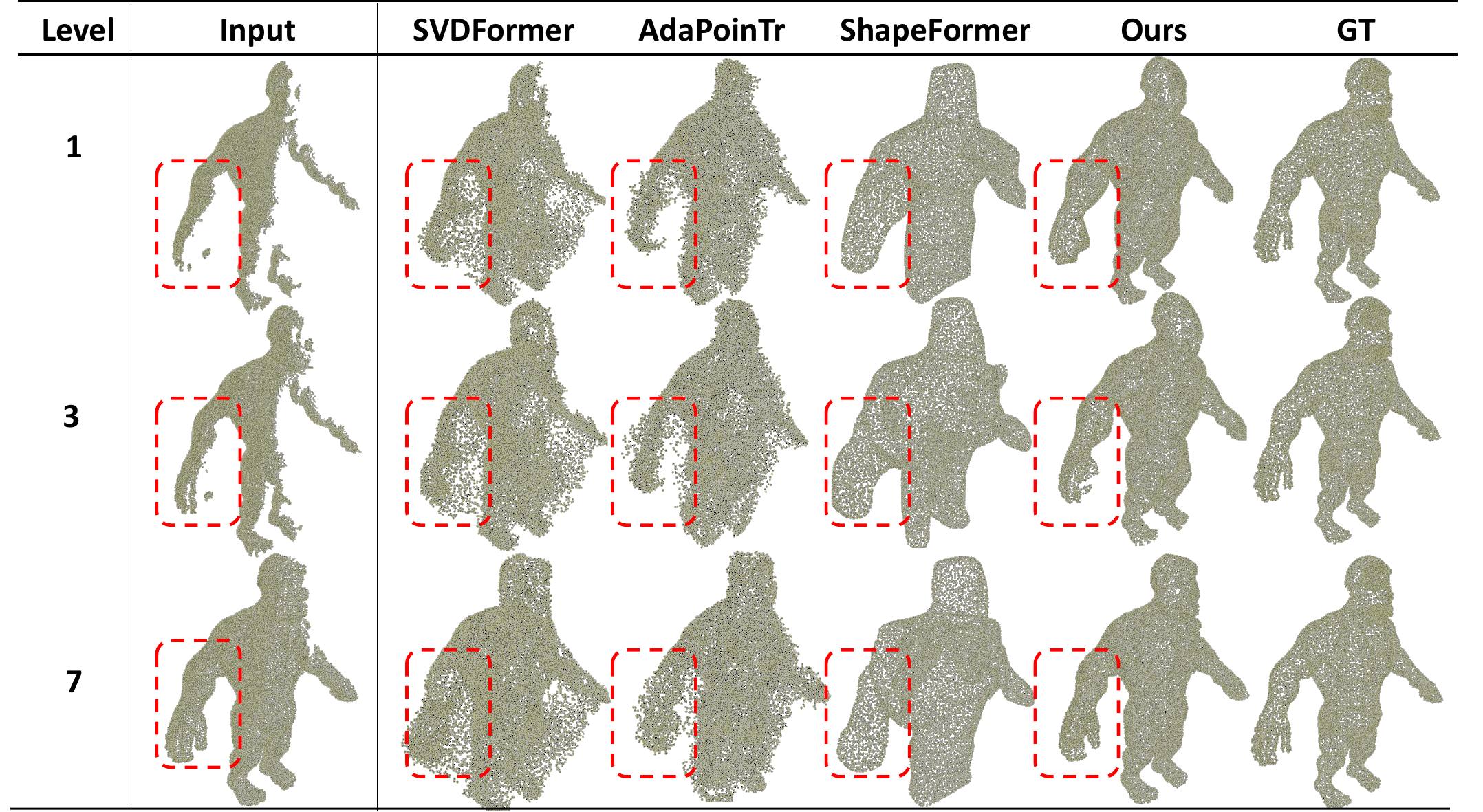}}
		}
		\vskip -0.1in
	\caption{\textcolor{black}{Qualitative comparisons under different incompleteness levels. The levels denote how many depth maps are used to construct the partial input.}}
	\label{pic:levels}
	\end{center}
	\vskip -0.3in
\end{figure*}

\textcolor{black}{In this section, we evaluate the performances of our method on partial input with different incompleteness levels. For convenience, we use synthetic objects from Sec.~\ref{sec:datasets} to construct evaluation sets with varying levels of incompleteness. Specifically, we initialize the first virtual camera at a pose of $elevation = 0$, $azimuth = -140$, and $fov \approx 80^\circ$. Additional virtual cameras are placed along the azimuth at 15° intervals. By merging 1, 3, and 7 consecutive depth maps, we generate partial point clouds with different levels of incompleteness. These data are used for comparison experiments. The qualitative and quantitative comparisons are presented in Fig.~\ref{pic:levels} and Table~\ref{tab:levels}, respectively.}
\textcolor{black}{We can see that more completed partial input constructed from more depth maps will bring finer details to the completed results. Our method consistently outperforms other methods in this setting.}

\subsection{Evaluation on Multi-modal metrics}
\textcolor{black}{Since our method relies on SDS guidance from Zero 1-to-3~\cite{liu2023zero}, it may produce different completion results with each optimization. To evaluate its performance under these variations, we assess our method using multi-modal metrics, including TMD, UHD, and MMD, following the approach in \citep{chou2023diffusion}.
We perform four repeated optimizations for both our method and SDS-complete on the in-domain categories of Redwood, as detailed in Sec.~\ref{sec:redwood}, to compute these metrics. The results are summarized in Table~\ref{tab:tmd}. Our method achieves superior performance on UHD and MMD metrics, further validating its effectiveness for 3D point cloud completion. Although it shows a lower TMD, which evaluates completion diversity, this actually reflects its steady convergence toward the ground truths—a positive attribute for the task of 3D point cloud completion.}
\begin{table}[ht]
  \caption{Quantitative comparisons on multi-modal metrics.}
		\begin{center}
  % \vskip -0.1 in
  \setlength\tabcolsep{1.2pt}
			\scalebox{0.9}{
				\begin{tabular}{c|cccccc}
\hline
Methods                       & Metrics & Table         & Exe-Chair     & Out-Chair     & Old-Chair      & Aver          \\ \hline
\multirow{3}{*}{SDS-Complete} & TMD $\uparrow$     & \textbf{1.26}          & \textbf{1.70}          & \textbf{1.18}          & \textbf{1.25}           & \textbf{1.35}          \\
                              & UHD $\downarrow$    & 9.31          & 10.39         & 10.63         & 16.67          & 11.75         \\
                              & MMD $\downarrow$    & \textbf{1.27} & 1.66          & 1.86          & 2.11           & 1.73          \\ \hline
\multirow{3}{*}{Ours}         & TMD $\uparrow$    & 0.57 & 0.53 & 0.42 & 0.65  & 0.54 \\
                              & UHD $\downarrow$    & \textbf{8.47} & \textbf{4.73} & \textbf{8.64} & \textbf{12.02} & \textbf{8.47} \\
                              & MMD $\downarrow$    & 1.47          & \textbf{1.04} & \textbf{1.28} & \textbf{1.42}  & \textbf{1.30} \\ \hline
\end{tabular}
			}
		\end{center}
		 % \vskip -0.1in
		\label{tab:tmd}
		\vskip -0.3in
	\end{table}
% \subsection{Comparisons on Multi-modal Metrics}

	\begin{table}[ht]
         % \vskip -0.1in
		\normalsize
		\setlength\tabcolsep{2.0pt}
  	\caption{Quantitative comparison on ShapeNet dataset. "Known category" and "Unknown category" denote categories included and not included in the training set of network-based methods, respectively.}
		\begin{center}
  % \vskip -0.2in
			\scalebox{0.9}{
				\begin{tabular}{ccccc}
\toprule
           & \multicolumn{2}{c}{Known category} & \multicolumn{2}{c}{Unknown category}           \\ \cmidrule{2-5}
Categories & Chair            & Table            & Pistol            & Tower              \\
Metrics    & CD/EMD           & CD/EMD           & CD/EMD            & CD/EMD               \\ \midrule
PoinTr     & \textbf{1.31}/2.64        & \textbf{0.74}/2.86        & 1.84/3.84         & 2.38/3.05         \\
SeedFormer & 1.39/2.77        & 0.80/2.17        & 1.79/3.91         & 1.95/3.24         \\
AdaPoinTr  & 1.45/2.54        & 0.74/1.58        & 2.28/4.15         & 1.99/3.30         \\
PointAttN  & 1.26/2.56        & 0.92/1.93        & 2.48/4.83         & 1.72/3.03       \\
SVDFormer  & 1.21/2.49        & 1.68/3.15        & 2.02/4.25         & 3.47/4.22       \\
Ours       & 1.38/\textbf{1.94}        & 1.08/\textbf{1.56}        & \textbf{1.09}/\textbf{1.63}         & \textbf{1.41}/\textbf{1.82}       \\ \bottomrule
\end{tabular}
			}
		\end{center}
		% \vskip -0.2in
		\label{quan_shapenet}
		\vskip -0.2in
	\end{table}
\begin{figure*}[t]
	%%\vskip -0.3in
	\begin{center}
		\scalebox{0.9}{
			\centerline{\includegraphics[width=\linewidth]{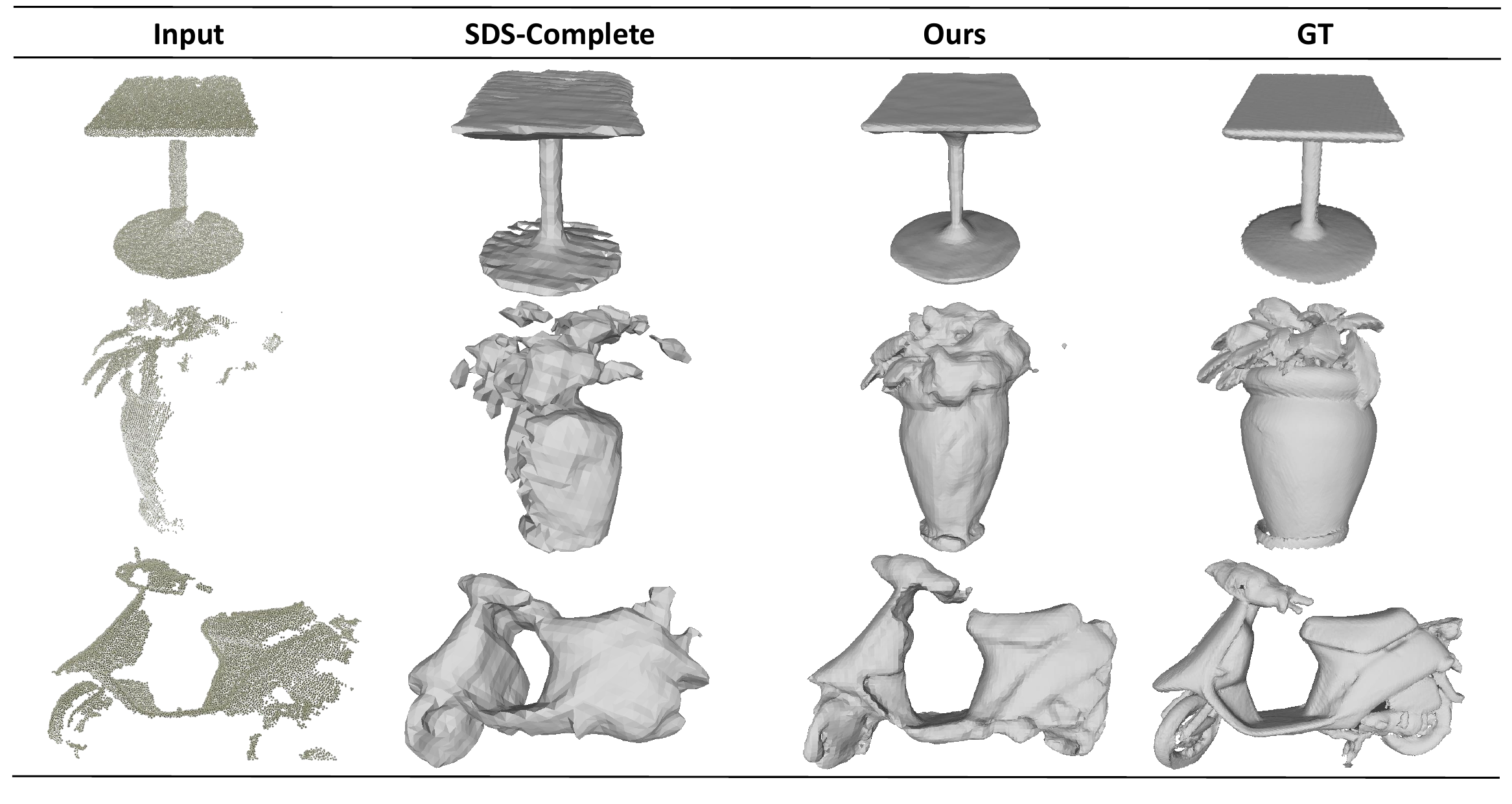}}
		}
		\vskip -0.2in
	\caption{Comparisons based on Meshes.}
	\label{pic:mesh}
	\end{center}
	\vskip -0.3in
\end{figure*}
\subsection{Comparisons based on Meshes}
\textcolor{black}{Our method potentially support the generation of 3D meshes due to the introducing of Grid Pulling module. As mentioned in Sec~\ref{sec:metho}, the Grip Pulling is proposed to re-sample uniform points from the non-uniformed point cloud $P_{surf}$, where a SDF fuction $g(\cdot)$ is introduced to fit the overall shape of $P_{surf}$ to do the resampling. Therefore, we can use Marching Cubes following NeuralPull~\cite{ma2020neural} to extract meshes from $g(\cdot)$.
The results are presented in Fig.~\ref{pic:mesh}. We can see that our method can also create more accurate mesh shapes than SDS-Complete~\cite{kasten2024point}.}
\subsection{Evaluation on ShapeNet}
\label{sec:sp}
\begin{figure*}[ht]
	%%\vskip -0.3in
	\begin{center}
		\scalebox{0.84}{
			\centerline{\includegraphics[width=\linewidth]{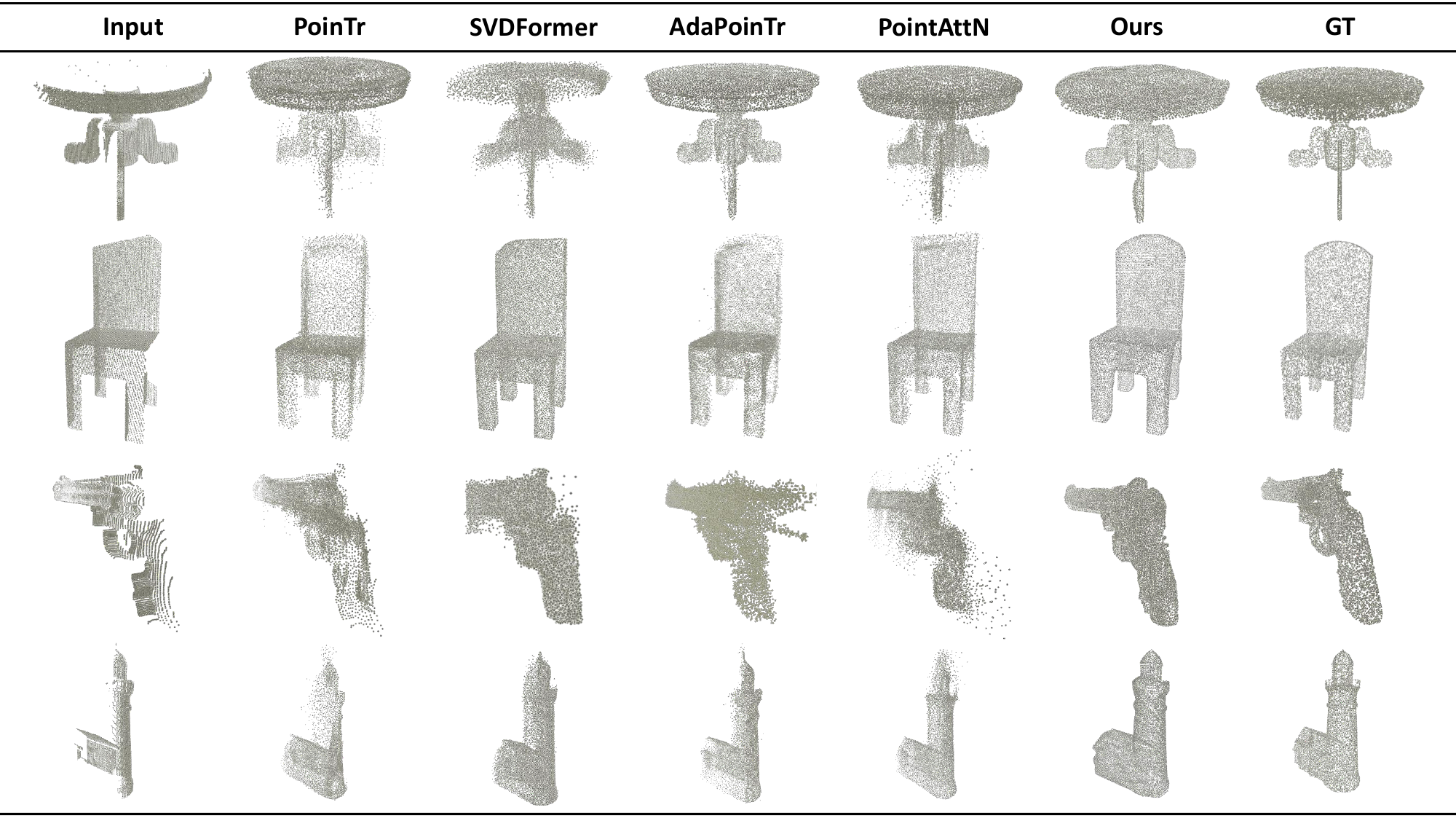}}
		}
		\vskip -0.1in
		\caption{Qualitative comparison on objects from ShapeNet~\citep{geiger2013vision} dataset.}
		\label{pic:sp_quali}
	\end{center}
	% \vskip -0.3in
\end{figure*}

% As most existing point cloud completion networks~\citep{yu2021pointr,zhou2022seedformer,yu2023adapointr,wang2024pointattn} are trained on point clouds from ShapeNet~\citep{chang2015shapenet} dataset, they perform quite well on these in-domain data. 
In this section, we further compare our methods with network-based methods on 16 common models from 4 different categories of ShapeNet dataset.
% The Chair and Table
The results are presented in Table~\ref{quan_shapenet} and Fig.~\ref{pic:sp_quali}. 
% We can see that our method performs relatively inferior to the network-based methods on the known categories.
% which have close distributions as their training data. 
Although our method performs slightly inferior to network-based methods on the known category objects, it surpasses other methods on the unknown category objects.
Please note that network-based methods use 3D ground truths from known categories for supervision during training, while our method does not introduce any training with such ground truths.
As a test-time point cloud completion method, the core contribution of our method is its generalizablity for point cloud objects from any category. This has been confirmed by experiments on multiple kinds of data including synthetic objects in Sec.~\ref{sec:objs}, Redwood dataset in Sec.~\ref{sec:redwood}, and ShapeNet in Sec.~\ref{sec:sp}. 
% such 3D supervision as a zero-shot method. 
% We also present qualitative comparisons in Fig.~\ref{pic:sp_quali}. Our method can always creates correct overall contours on different categories, which validates the effectiveness of our method as a zero-shot method.
% , though it also produces 
% It would be interesting to observe the performance of our zero-shot method on these data to 
 % "Known category" denotes the categories are included in the training set of network-based completion methods, while "Unknown category" means the categories are unseen.

\subsection{Evaluation on LiDAR Points}
\label{sec:kitti}
\begin{figure*}[ht]
	%%\vskip -0.3in
	\begin{center}
		\scalebox{0.84}{
			\centerline{\includegraphics[width=\linewidth]{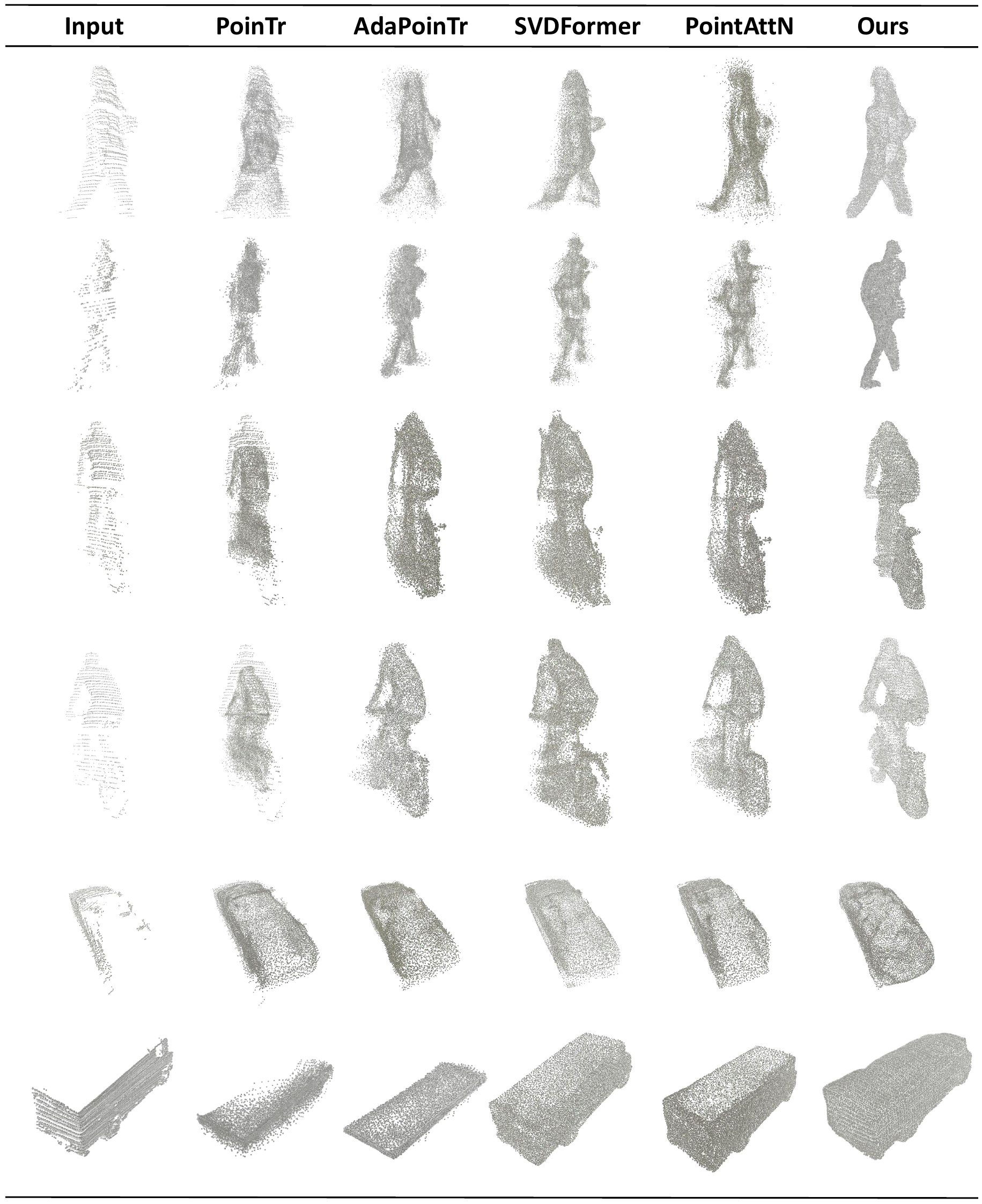}}
		}
		\vskip -0.1in
		\caption{Comparison on Kitti~\citep{geiger2013vision} dataset.}
		\label{pic:kitti_supp}
	\end{center}
	% \vskip -0.3in
\end{figure*}

As discussed in Sec.~\ref{sec:metho}, we render the reference image $I_{in}$ from the incomplete point cloud $P_{in}$, under the estimated camera pose $V_p$. This operation means that we actually observe the point cloud from a pinhole camera model, which may be closer to point clouds from depth scanners, such as Redwood dataset~\citep{choi2016large,kasten2024point}. 
To validate the effectiveness of our method across different sensor types, we conduct a comparison using point clouds from the Kitti dataset~\citep{geiger2013vision}, which are acquired with LiDAR sensors. Point clouds from Pedestrian, Cyclist, Car, and Truck are adopted for evaluation.
Since ground truth data are unavailable for these point clouds, we mainly present qualitative comparison in Fig.~\ref{pic:kitti_supp}. Notably, our method demonstrates the ability for reasonable completion even with LiDAR-derived point clouds.

% As MMD metric~\citep{yuan2018pcn} measures the average distance between the completed point clouds and their nearest neighbors in ShapeNet dataset~\citep{chang2015shapenet}, it is actually more conductive to the network-based methods pre-trained on ShapeNet. Besides, Pedestrian and Cyclist does not exist in ShapeNet. Their MMD cannot be calculated.

% The MMD metric, as described in~\citep{yuan2018pcn}, calculates the average distance between completed point clouds and their nearest neighbors in the ShapeNet dataset~\citep{chang2015shapenet}. However, this metric is more favorable to network-based methods that have been pre-trained on ShapeNet. Additionally, categories such as Pedestrians and Cyclists are absent in ShapeNet, making it impossible to calculate their MMD.
% Therefore, we use Fidelity Distance (FD)~\citep{yuan2018pcn} as a supplement quantitative metric in Table~\ref{quan_kitti}.
% 	\begin{table}[h]
%          \vskip -0.1in
% 		\normalsize
% 		\setlength\tabcolsep{2.0pt}
% 		%\renewcommand\arraystretch{1.5}
%   	\caption{Quantitative comparison on objects from Kitti.}
% 		\begin{center}
%   % \vskip -0.2in
% 			\scalebox{0.9}{
% 				\begin{tabular}{cccccc}
%     \toprule
%    & PoinTr & AdaPointr & Seedformer & PointAttN & Ours \\
%    \midrule
% FD & 0.71   & 3.02      & 0.62       & 2.63      & \textbf{0.41} \\
% \bottomrule
% \end{tabular}
% 			}
% 		\end{center}
% 		% \vskip -0.2in
% 		\label{quan_kitti}
% 		% \vskip -0.3in
% 	\end{table}
% We can see our method gets smallest FD, which means it preserves details from partial point clouds well.

%% file: iclr2025_conference.bbl
\begin{thebibliography}{47}
\providecommand{\natexlab}[1]{#1}
\providecommand{\url}[1]{\texttt{#1}}
\expandafter\ifx\csname urlstyle\endcsname\relax
  \providecommand{\doi}[1]{doi: #1}\else
  \providecommand{\doi}{doi: \begingroup \urlstyle{rm}\Url}\fi

\bibitem[Breckon \& Fisher(2005)Breckon and Fisher]{breckon2005amodal}
Toby~P Breckon and Robert~B Fisher.
\newblock Amodal volume completion: 3d visual completion.
\newblock \emph{Computer Vision and Image Understanding}, 99\penalty0 (3):\penalty0 499--526, 2005.

\bibitem[Cadena et~al.(2016)Cadena, Carlone, Carrillo, Latif, Scaramuzza, Neira, Reid, and Leonard]{cadena2016past}
Cesar Cadena, Luca Carlone, Henry Carrillo, Yasir Latif, Davide Scaramuzza, Jos{\'e} Neira, Ian Reid, and John~J Leonard.
\newblock Past, present, and future of simultaneous localization and mapping: Toward the robust-perception age.
\newblock \emph{IEEE Transactions on robotics}, 32\penalty0 (6):\penalty0 1309--1332, 2016.

\bibitem[Chang et~al.(2015)Chang, Funkhouser, Guibas, Hanrahan, Huang, Li, Savarese, Savva, Song, Su, et~al.]{chang2015shapenet}
Angel~X Chang, Thomas Funkhouser, Leonidas Guibas, Pat Hanrahan, Qixing Huang, Zimo Li, Silvio Savarese, Manolis Savva, Shuran Song, Hao Su, et~al.
\newblock Shapenet: An information-rich 3d model repository.
\newblock \emph{arXiv preprint arXiv:1512.03012}, 2015.

\bibitem[Choi et~al.(2016)Choi, Zhou, Miller, and Koltun]{choi2016large}
Sungjoon Choi, Qian-Yi Zhou, Stephen Miller, and Vladlen Koltun.
\newblock A large dataset of object scans.
\newblock \emph{arXiv preprint arXiv:1602.02481}, 2016.

\bibitem[Chou et~al.(2023)Chou, Bahat, and Heide]{chou2023diffusion}
Gene Chou, Yuval Bahat, and Felix Heide.
\newblock Diffusion-sdf: Conditional generative modeling of signed distance functions.
\newblock In \emph{Proceedings of the IEEE/CVF international conference on computer vision}, pp.\  2262--2272, 2023.

\bibitem[Chu et~al.(2023)Chu, Xie, Mo, Li, Nie{\ss}ner, Fu, and Jia]{chu2024diffcomplete}
Ruihang Chu, Enze Xie, Shentong Mo, Zhenguo Li, Matthias Nie{\ss}ner, Chi-Wing Fu, and Jiaya Jia.
\newblock Diffcomplete: Diffusion-based generative 3d shape completion.
\newblock \emph{Advances in Neural Information Processing Systems}, 2023.

\bibitem[DeCarlo et~al.(2003)DeCarlo, Finkelstein, Rusinkiewicz, and Santella]{DBLP:journals/tog/DeCarloFRS03}
Douglas DeCarlo, Adam Finkelstein, Szymon Rusinkiewicz, and Anthony Santella.
\newblock Suggestive contours for conveying shape.
\newblock \emph{{ACM} Trans. Graph.}, 22\penalty0 (3):\penalty0 848--855, 2003.
\newblock \doi{10.1145/882262.882354}.
\newblock URL \url{https://doi.org/10.1145/882262.882354}.

\bibitem[Fan et~al.(2017)Fan, Su, and Guibas]{fan2017point}
Haoqiang Fan, Hao Su, and Leonidas~J Guibas.
\newblock A point set generation network for 3d object reconstruction from a single image.
\newblock In \emph{Proceedings of the IEEE conference on computer vision and pattern recognition}, pp.\  605--613, 2017.

\bibitem[Geiger et~al.(2013)Geiger, Lenz, Stiller, and Urtasun]{geiger2013vision}
Andreas Geiger, Philip Lenz, Christoph Stiller, and Raquel Urtasun.
\newblock Vision meets robotics: The kitti dataset.
\newblock \emph{The International Journal of Robotics Research}, 32\penalty0 (11):\penalty0 1231--1237, 2013.

\bibitem[Ho et~al.(2020)Ho, Jain, and Abbeel]{ho2020denoising}
Jonathan Ho, Ajay Jain, and Pieter Abbeel.
\newblock Denoising diffusion probabilistic models.
\newblock \emph{Advances in neural information processing systems}, 33:\penalty0 6840--6851, 2020.

\bibitem[Huang et~al.(2021)Huang, Zou, Cui, Yang, Wang, Zhao, Zhang, Yuan, Xu, and Liu]{huang2021rfnet}
Tianxin Huang, Hao Zou, Jinhao Cui, Xuemeng Yang, Mengmeng Wang, Xiangrui Zhao, Jiangning Zhang, Yi~Yuan, Yifan Xu, and Yong Liu.
\newblock Rfnet: Recurrent forward network for dense point cloud completion.
\newblock In \emph{Proceedings of the IEEE/CVF international conference on computer vision}, pp.\  12508--12517, 2021.

\bibitem[Huang et~al.(2022)Huang, Zhang, Chen, Ding, Tai, Zhang, Wang, and Liu]{huang20223qnet}
Tianxin Huang, Jiangning Zhang, Jun Chen, Zhonggan Ding, Ying Tai, Zhenyu Zhang, Chengjie Wang, and Yong Liu.
\newblock 3qnet: 3d point cloud geometry quantization compression network.
\newblock \emph{ACM Transactions on Graphics (TOG)}, 41\penalty0 (6):\penalty0 1--13, 2022.

\bibitem[Huang et~al.(2020)Huang, Yu, Xu, Ni, and Le]{huang2020pf}
Zitian Huang, Yikuan Yu, Jiawen Xu, Feng Ni, and Xinyi Le.
\newblock Pf-net: Point fractal network for 3d point cloud completion.
\newblock In \emph{Proceedings of the IEEE/CVF Conference on Computer Vision and Pattern Recognition}, pp.\  7662--7670, 2020.

\bibitem[Kasten et~al.(2024)Kasten, Rahamim, and Chechik]{kasten2024point}
Yoni Kasten, Ohad Rahamim, and Gal Chechik.
\newblock Point cloud completion with pretrained text-to-image diffusion models.
\newblock \emph{Advances in Neural Information Processing Systems}, 36, 2024.

\bibitem[Kerbl et~al.(2023)Kerbl, Kopanas, Leimk{\"u}hler, and Drettakis]{kerbl20233d}
Bernhard Kerbl, Georgios Kopanas, Thomas Leimk{\"u}hler, and George Drettakis.
\newblock 3d gaussian splatting for real-time radiance field rendering.
\newblock \emph{ACM Transactions on Graphics}, 42\penalty0 (4), 2023.

\bibitem[Krishnamurthy \& Levoy(1996)Krishnamurthy and Levoy]{DBLP:conf/siggraph/KrishnamurthyL96}
Venkat Krishnamurthy and Marc Levoy.
\newblock Fitting smooth surfaces to dense polygon meshes.
\newblock In John Fujii (ed.), \emph{Proceedings of the 23rd Annual Conference on Computer Graphics and Interactive Techniques, {SIGGRAPH} 1996, New Orleans, LA, USA, August 4-9, 1996}, pp.\  313--324. {ACM}, 1996.
\newblock \doi{10.1145/237170.237270}.
\newblock URL \url{https://doi.org/10.1145/237170.237270}.

\bibitem[Lehar(1999)]{lehar1999gestalt}
Steven Lehar.
\newblock Gestalt isomorphism and the quantification of spatial perception.
\newblock \emph{Gestalt theory}, 21:\penalty0 122--139, 1999.

\bibitem[Li et~al.(2023)Li, Gao, Tan, and Wei]{li2023proxyformer}
Shanshan Li, Pan Gao, Xiaoyang Tan, and Mingqiang Wei.
\newblock Proxyformer: Proxy alignment assisted point cloud completion with missing part sensitive transformer.
\newblock In \emph{Proceedings of the IEEE/CVF conference on computer vision and pattern recognition}, pp.\  9466--9475, 2023.

\bibitem[Lipman et~al.(2008)Lipman, Levin, and Cohen{-}Or]{DBLP:journals/tog/LipmanLC08}
Yaron Lipman, David Levin, and Daniel Cohen{-}Or.
\newblock Green coordinates.
\newblock \emph{{ACM} Trans. Graph.}, 27\penalty0 (3):\penalty0 78, 2008.
\newblock \doi{10.1145/1360612.1360677}.
\newblock URL \url{https://doi.org/10.1145/1360612.1360677}.

\bibitem[Liu et~al.(2023)Liu, Wu, Van~Hoorick, Tokmakov, Zakharov, and Vondrick]{liu2023zero}
Ruoshi Liu, Rundi Wu, Basile Van~Hoorick, Pavel Tokmakov, Sergey Zakharov, and Carl Vondrick.
\newblock Zero-1-to-3: Zero-shot one image to 3d object.
\newblock In \emph{Proceedings of the IEEE/CVF International Conference on Computer Vision}, pp.\  9298--9309, 2023.

\bibitem[Ma et~al.(2020)Ma, Han, Liu, and Zwicker]{ma2020neural}
Baorui Ma, Zhizhong Han, Yu-Shen Liu, and Matthias Zwicker.
\newblock Neural-pull: Learning signed distance functions from point clouds by learning to pull space onto surfaces.
\newblock \emph{arXiv preprint arXiv:2011.13495}, 2020.

\bibitem[Michel et~al.(2022)Michel, Bar-On, Liu, Benaim, and Hanocka]{michel2022text2mesh}
Oscar Michel, Roi Bar-On, Richard Liu, Sagie Benaim, and Rana Hanocka.
\newblock Text2mesh: Text-driven neural stylization for meshes.
\newblock In \emph{Proceedings of the IEEE/CVF Conference on Computer Vision and Pattern Recognition}, pp.\  13492--13502, 2022.

\bibitem[Mohammad~Khalid et~al.(2022)Mohammad~Khalid, Xie, Belilovsky, and Popa]{mohammad2022clip}
Nasir Mohammad~Khalid, Tianhao Xie, Eugene Belilovsky, and Tiberiu Popa.
\newblock Clip-mesh: Generating textured meshes from text using pretrained image-text models.
\newblock In \emph{SIGGRAPH Asia 2022 conference papers}, pp.\  1--8, 2022.

\bibitem[Poole et~al.(2022)Poole, Jain, Barron, and Mildenhall]{poole2022dreamfusion}
Ben Poole, Ajay Jain, Jonathan~T Barron, and Ben Mildenhall.
\newblock Dreamfusion: Text-to-3d using 2d diffusion.
\newblock \emph{arXiv preprint arXiv:2209.14988}, 2022.

\bibitem[Praun et~al.(2000)Praun, Finkelstein, and Hoppe]{DBLP:conf/siggraph/PraunFH00}
Emil Praun, Adam Finkelstein, and Hugues Hoppe.
\newblock Lapped textures.
\newblock In Judith~R. Brown and Kurt Akeley (eds.), \emph{Proceedings of the 27th Annual Conference on Computer Graphics and Interactive Techniques, {SIGGRAPH} 2000, New Orleans, LA, USA, July 23-28, 2000}, pp.\  465--470. {ACM}, 2000.
\newblock \doi{10.1145/344779.344987}.
\newblock URL \url{https://doi.org/10.1145/344779.344987}.

\bibitem[Reddy et~al.(2018)Reddy, Vo, and Narasimhan]{reddy2018carfusion}
N~Dinesh Reddy, Minh Vo, and Srinivasa~G Narasimhan.
\newblock Carfusion: Combining point tracking and part detection for dynamic 3d reconstruction of vehicles.
\newblock In \emph{Proceedings of the IEEE conference on computer vision and pattern recognition}, pp.\  1906--1915, 2018.

\bibitem[Rombach et~al.(2022)Rombach, Blattmann, Lorenz, Esser, and Ommer]{rombach2022high}
Robin Rombach, Andreas Blattmann, Dominik Lorenz, Patrick Esser, and Bj{\"o}rn Ommer.
\newblock High-resolution image synthesis with latent diffusion models.
\newblock In \emph{Proceedings of the IEEE/CVF conference on computer vision and pattern recognition}, pp.\  10684--10695, 2022.

\bibitem[Saharia et~al.(2022)Saharia, Chan, Saxena, Li, Whang, Denton, Ghasemipour, Gontijo~Lopes, Karagol~Ayan, Salimans, et~al.]{saharia2022photorealistic}
Chitwan Saharia, William Chan, Saurabh Saxena, Lala Li, Jay Whang, Emily~L Denton, Kamyar Ghasemipour, Raphael Gontijo~Lopes, Burcu Karagol~Ayan, Tim Salimans, et~al.
\newblock Photorealistic text-to-image diffusion models with deep language understanding.
\newblock \emph{Advances in Neural Information Processing Systems}, 35:\penalty0 36479--36494, 2022.

\bibitem[Tang et~al.(2023)Tang, Ren, Zhou, Liu, and Zeng]{tang2023dreamgaussian}
Jiaxiang Tang, Jiawei Ren, Hang Zhou, Ziwei Liu, and Gang Zeng.
\newblock Dreamgaussian: Generative gaussian splatting for efficient 3d content creation.
\newblock \emph{arXiv preprint arXiv:2309.16653}, 2023.

\bibitem[Tchapmi et~al.(2019)Tchapmi, Kosaraju, Rezatofighi, Reid, and Savarese]{tchapmi2019topnet}
Lyne~P Tchapmi, Vineet Kosaraju, Hamid Rezatofighi, Ian Reid, and Silvio Savarese.
\newblock Topnet: Structural point cloud decoder.
\newblock In \emph{Proceedings of the IEEE Conference on Computer Vision and Pattern Recognition}, pp.\  383--392, 2019.

\bibitem[Wang et~al.(2023)Wang, Du, Li, Yeh, and Shakhnarovich]{wang2023score}
Haochen Wang, Xiaodan Du, Jiahao Li, Raymond~A Yeh, and Greg Shakhnarovich.
\newblock Score jacobian chaining: Lifting pretrained 2d diffusion models for 3d generation.
\newblock In \emph{Proceedings of the IEEE/CVF Conference on Computer Vision and Pattern Recognition}, pp.\  12619--12629, 2023.

\bibitem[Wang et~al.(2024)Wang, Cui, Guo, Li, Liu, and Shen]{wang2024pointattn}
Jun Wang, Ying Cui, Dongyan Guo, Junxia Li, Qingshan Liu, and Chunhua Shen.
\newblock Pointattn: You only need attention for point cloud completion.
\newblock In \emph{Proceedings of the AAAI Conference on artificial intelligence}, pp.\  5472--5480, 2024.

\bibitem[Wang et~al.(2020)Wang, Ang~Jr, and Lee]{wang2020cascaded}
Xiaogang Wang, Marcelo~H Ang~Jr, and Gim~Hee Lee.
\newblock Cascaded refinement network for point cloud completion.
\newblock In \emph{Proceedings of the IEEE/CVF Conference on Computer Vision and Pattern Recognition}, pp.\  790--799, 2020.

\bibitem[Wen et~al.(2021)Wen, Xiang, Han, Cao, Wan, Zheng, and Liu]{wen2021pmp}
Xin Wen, Peng Xiang, Zhizhong Han, Yan-Pei Cao, Pengfei Wan, Wen Zheng, and Yu-Shen Liu.
\newblock Pmp-net: Point cloud completion by learning multi-step point moving paths.
\newblock In \emph{Proceedings of the IEEE/CVF conference on computer vision and pattern recognition}, pp.\  7443--7452, 2021.

\bibitem[Xiang et~al.(2022)Xiang, Wen, Liu, Cao, Wan, Zheng, and Han]{xiang2022snowflake}
Peng Xiang, Xin Wen, Yu-Shen Liu, Yan-Pei Cao, Pengfei Wan, Wen Zheng, and Zhizhong Han.
\newblock Snowflake point deconvolution for point cloud completion and generation with skip-transformer.
\newblock \emph{IEEE Transactions on Pattern Analysis and Machine Intelligence}, 45\penalty0 (5):\penalty0 6320--6338, 2022.

\bibitem[Xie et~al.(2020)Xie, Yao, Zhou, Mao, Zhang, and Sun]{xie2020grnet}
Haozhe Xie, Hongxun Yao, Shangchen Zhou, Jiageng Mao, Shengping Zhang, and Wenxiu Sun.
\newblock Grnet: Gridding residual network for dense point cloud completion.
\newblock \emph{arXiv preprint arXiv:2006.03761}, 2020.

\bibitem[Yan et~al.(2022)Yan, Lin, Mitra, Lischinski, Cohen-Or, and Huang]{yan2022shapeformer}
Xingguang Yan, Liqiang Lin, Niloy~J Mitra, Dani Lischinski, Daniel Cohen-Or, and Hui Huang.
\newblock Shapeformer: Transformer-based shape completion via sparse representation.
\newblock In \emph{Proceedings of the IEEE/CVF Conference on Computer Vision and Pattern Recognition}, pp.\  6239--6249, 2022.

\bibitem[Yariv et~al.(2021)Yariv, Gu, Kasten, and Lipman]{yariv2021volume}
Lior Yariv, Jiatao Gu, Yoni Kasten, and Yaron Lipman.
\newblock Volume rendering of neural implicit surfaces.
\newblock \emph{Advances in Neural Information Processing Systems}, 34:\penalty0 4805--4815, 2021.

\bibitem[Yu et~al.(2021)Yu, Rao, Wang, Liu, Lu, and Zhou]{yu2021pointr}
Xumin Yu, Yongming Rao, Ziyi Wang, Zuyan Liu, Jiwen Lu, and Jie Zhou.
\newblock Pointr: Diverse point cloud completion with geometry-aware transformers.
\newblock In \emph{Proceedings of the IEEE/CVF international conference on computer vision}, pp.\  12498--12507, 2021.

\bibitem[Yu et~al.(2023)Yu, Rao, Wang, Lu, and Zhou]{yu2023adapointr}
Xumin Yu, Yongming Rao, Ziyi Wang, Jiwen Lu, and Jie Zhou.
\newblock Adapointr: Diverse point cloud completion with adaptive geometry-aware transformers.
\newblock \emph{arXiv preprint arXiv:2301.04545}, 2023.

\bibitem[Yuan et~al.(2018)Yuan, Khot, Held, Mertz, and Hebert]{yuan2018pcn}
Wentao Yuan, Tejas Khot, David Held, Christoph Mertz, and Martial Hebert.
\newblock Pcn: Point completion network.
\newblock In \emph{2018 International Conference on 3D Vision (3DV)}, pp.\  728--737. IEEE, 2018.

\bibitem[Zhang et~al.(2023)Zhang, Rao, and Agrawala]{zhang2023adding}
Lvmin Zhang, Anyi Rao, and Maneesh Agrawala.
\newblock Adding conditional control to text-to-image diffusion models.
\newblock In \emph{Proceedings of the IEEE/CVF International Conference on Computer Vision}, pp.\  3836--3847, 2023.

\bibitem[Zhang et~al.(2020)Zhang, Yan, and Xiao]{zhang2020detail}
Wenxiao Zhang, Qingan Yan, and Chunxia Xiao.
\newblock Detail preserved point cloud completion via separated feature aggregation.
\newblock \emph{arXiv preprint arXiv:2007.02374}, 2020.

\bibitem[Zhao et~al.(2021)Zhao, Jiang, Jia, Torr, and Koltun]{zhao2021point}
Hengshuang Zhao, Li~Jiang, Jiaya Jia, Philip~HS Torr, and Vladlen Koltun.
\newblock Point transformer.
\newblock In \emph{Proceedings of the IEEE/CVF international conference on computer vision}, pp.\  16259--16268, 2021.

\bibitem[Zhou et~al.(2022)Zhou, Cao, Chu, Zhu, Lu, Tai, and Wang]{zhou2022seedformer}
Haoran Zhou, Yun Cao, Wenqing Chu, Junwei Zhu, Tong Lu, Ying Tai, and Chengjie Wang.
\newblock Seedformer: Patch seeds based point cloud completion with upsample transformer.
\newblock In \emph{European conference on computer vision}, pp.\  416--432. Springer, 2022.

\bibitem[Zhou et~al.(2018)Zhou, Park, and Koltun]{zhou2018open3d}
Qian-Yi Zhou, Jaesik Park, and Vladlen Koltun.
\newblock Open3d: A modern library for 3d data processing.
\newblock \emph{arXiv preprint arXiv:1801.09847}, 2018.

\bibitem[Zhu et~al.(2023)Zhu, Chen, He, Wang, Qin, and Wei]{Zhu_2023_ICCV}
Zhe Zhu, Honghua Chen, Xing He, Weiming Wang, Jing Qin, and Mingqiang Wei.
\newblock Svdformer: Complementing point cloud via self-view augmentation and self-structure dual-generator.
\newblock In \emph{Proceedings of the IEEE/CVF International Conference on Computer Vision (ICCV)}, pp.\  14508--14518, October 2023.

\end{thebibliography}
